\documentclass[Afour,sageh,times]{sagej}

\usepackage[ruled,vlined,linesnumbered]{algorithm2e}
\usepackage{mathrsfs}
\usepackage{graphicx}
\usepackage{amsfonts}
\usepackage{amsmath}
\usepackage{amssymb}
\usepackage{array}
\usepackage{subfig}
\usepackage{verbatim}
\usepackage[svgnames,dvipsnames]{xcolor}
\usepackage{siunitx}

\usepackage{balance}
\usepackage{tabularx}

\usepackage{floatrow}

\usepackage{nicefrac}
\usepackage{moreverb,url}
\usepackage[colorlinks,bookmarksopen,bookmarksnumbered,citecolor=red,urlcolor=red]{hyperref}

\usepackage{amsthm}	%

\usepackage{etoolbox}

\newtheorem{Lem}{Lemma}

\newtheorem{assumption}{Assumption}
\newtheorem{Remark}{Remark}

\newenvironment{proof}[1][Proof.]{\begin{trivlist}
\item[\hskip \labelsep {\bfseries #1}]}{\end{trivlist}}

\newtoggle{finalpaper}
\togglefalse{finalpaper}

\allowdisplaybreaks[1]

\newcommand{\pr}[1]{\textbf{#1:} }

\newcommand{\rev}[1]{{\color{black}#1}}
\newcommand{\revf}[1]{{\color{black}#1}}

\usepackage{enumitem}

\usepackage{mathptmx}       %
\usepackage{helvet}         %
\usepackage{courier}        %
\usepackage{type1cm}        %
\usepackage{makeidx}         %

\makeindex             %

\newcommand{\flag}{Arxiv} %
\newcommand{\ifISRR}{\equal{\flag}{ISRR}} %
\newcommand{\ifArxiv}{\equal{\flag}{Arxiv}}

\newcommand{\ie}{\textit{i}.\textit{e}.,~}
\newcommand{\eg}{\textit{e}.\textit{g}.,~}

\DeclareSIUnit{\mph}{mph}

\def\volumeyear{2018}

\newcommand\BibTeX{{\rmfamily B\kern-.05em \textsc{i\kern-.025em b}\kern-.08em
T\kern-.1667em\lower.7ex\hbox{E}\kern-.125emX}}

\begin{document}

\title{Confidence-rich grid mapping}

\runninghead{Agha-mohammadi, Heiden, Hausman, Sukhatme}

\author{Ali-akbar Agha-mohammadi\affilnum{1}, Eric Heiden\affilnum{2}, Karol Hausman\affilnum{3}, Gaurav Sukhatme\affilnum{2}}

\affiliation{\affilnum{1}Jet Propulsion Laboratory, California Institute of Technology, Pasadena, CA 91109, USA\\
\affilnum{2}Department of Computer Science, University of Southern California, Los Angeles, CA 90089, USA\\
\affilnum{3}Google AI Research, Mountain View, CA 94043, USA}
        
\corrauth{Ali-akbar Agha-mohammadi,
        Jet Propulsion Laboratory,
        California Institute of Technology,
        Pasadena, CA 91109, USA}

\email{aliakbar.aghamohammadi@jpl.nasa.gov}

\begin{abstract}
\rev{Representing the environment is a fundamental task in enabling robots to act autonomously in unknown environments. In this work, we present confidence-rich mapping (CRM), a new algorithm for spatial grid-based mapping of the 3D environment. CRM augments the occupancy level at each voxel by its confidence value. By explicitly storing and evolving confidence values using the CRM filter, CRM extends traditional grid mapping in three ways: first, it partially maintains the probabilistic dependence among voxels. Second, it relaxes the need for hand-engineering an inverse sensor model and proposes the concept of sensor cause model that can be derived in a principled manner from the forward sensor model. Third, and most importantly, it provides consistent confidence values over the occupancy estimation that can be reliably used in collision risk evaluation and motion planning. CRM runs online and enables mapping environments where voxels might be partially occupied. We demonstrate the performance of the method on various datasets and environments in simulation and on physical systems. We show in real-world experiments that, in addition to achieving maps that are more accurate than traditional methods, the proposed filtering scheme demonstrates a much higher level of consistency between its error and the reported confidence, hence, enabling a more reliable collision risk evaluation for motion planning.
}
\end{abstract}


\maketitle

\section{Introduction} \label{subsec:intro}
Consider a quadrotor equipped with a forward-facing stereo camera flying in an obstacle-laden environment tasked to reach a goal.
In order to ensure the safety of the system and avoid collisions, the robot needs to create a representation of obstacles, which we refer to as the map, and incorporate it in the planning framework. 
Due to the noise caused by using imperfect sensors and models, the robot requires a probabilistic representation of the map that is able to capture the uncertainty of the environment.
In order to plan trajectories using such a probabilistic map, the planner needs to be aware of not only the occupancy value estimates but also of how much these values can be trusted. 
In particular, this trust (or confidence) is important when considering sensors with high noise, such as stereo cameras with small baselines.
This paper presents an algorithm that creates a map which, in addition to the most likely occupancy values, encodes the confidence (trust) associated with these values.

Grid-based structures are among the most common representations of the environment when dealing with range sensors~\citep{wurm2010octomap}. 
Typically, each grid voxel contains binary information indicating whether the voxel is free or occupied. 
In a slightly richer format, each voxel contains the probability of being occupied. 		
In the main body of literature, occupancy grids are used to store binary occupancies updated by the Log-Odds method \citep{Thrun2005,thrun2002robotic}, which is discussed in detail in Sec.~\ref{sec:background}. 
Even though the Log-Odds-based occupancy grids have enjoyed success in a variety of applications, these methods, especially when coping with noisy sensors such as sonar and stereo cameras, suffer from three main issues:
\begin{enumerate}[label=\Alph*)] %
	\item The occupancy of each voxel is updated independently of the rest of the map. 
	This is a well-known problem \citep{Thrun2005} which has been shown to lead to conflicts between the map and measurement data. In particular, when the sensor is noisy or has a large field of view, there is a clear coupling between voxels that fall into the field of view of the sensor.
	\item The Log-Odds methods rely on the ``inverse sensor model'' (ISM), which needs to be hand-engineered for each sensor and a given environment.
	\item In order to represent the voxel occupancy, each voxel stores a single number. As a result, there is no consistent confidence or trust value to help the planner reason about the reliability of the estimated occupancy.
\end{enumerate}

\noindent
In this paper, we propose a method that partially relaxes these assumptions, generates more accurate maps, and provides a more consistent filtering mechanism than prior approaches. 
The highlights and contributions of this work are as follows:
\begin{enumerate}%
	\item The main assumption in traditional occupancy grid mapping is (partially) relaxed. We take into account the interdependence between voxels within the same measurement cone at every \rev{time} step. Further, the proposed method relaxes the binary assumption on the occupancy level and is able to cope with maps where each voxel is only partially occupied by obstacles.
	\item We replace the ad-hoc inverse sensor model by a novel ``sensor cause model'' which is computed based on the forward sensor model in a principled manner.
	\item In addition to the most likely occupancy value for each voxel, the proposed map representation contains confidence values (\eg variance) of voxel occupancies. The confidence information is crucial for planning over grid maps. We incorporate the sensor model and its uncertainties in characterizing the map accuracy.
	\item While the majority of approaches that relax the voxel-independence assumption are batch methods, %
	our method does not require logging of the data in an offline phase. Instead, the map can be updated online as the sensory data are received.
\end{enumerate}

\noindent
In our experiments, our method achieves an improvement over traditional approaches of up to $30\%$ in absolute error and up to two orders of magnitudes better variance consistency, according to the proposed inconsistency measure.
We believe that the map representation introduced here and the probabilistic map-update algorithm provide a significant step towards uncertainty-aware safe planning, a crucial component to enabling fast navigation in uncertain environments.

The current paper draws on earlier work published in a conference paper~\citep{agha2017crm}. Here, we provide a more detailed theoretical derivation of the confidence-rich grid mapping approach. Additionally, we present new results on real-robot datasets and widely-used benchmarks as well as our own dataset captured in a challenging obstacle-rich environment using physical robots equipped with stereo cameras.

\section{Related Work}
The first application of occupancy grids in robotics dates back to \citet{moravec1988sensor} and \citet{Elfes1989occupancy} and has since been widely used in robotics. \citet{Thrun2005}, \citet{stachniss2009_book}, and \citet{thrun2002robotic} discuss many variants of these methods. 

Grid-based maps have been constructed using a variety of ranging sensors, that include stereo cameras \citep{konolige2008outdoor}, sonars \citep{yamauchi1997frontier}, laser range finders \citep{thrun1998learning}, and fusion of thereof~\citep{moravec1988sensor}. Their structure has been extended to achieve more memory-efficient maps \citep{wurm2010octomap}. 
\rev{Space carving approaches~\citep{kutulakos2000shape,martin1983volumetric} often use voxel-based representations to reconstruct 3D shapes of scenes and objects from multiple photographs obtained from different viewpoints.}
There have also been methods that extend grid-based mapping to store richer forms of data, including the distance to obstacle surfaces \citep{newcombe2011kinectfusion, oleynikova2017voxblox}, reflective properties of the environment \citep{howard1996generating}, as well as color and texture \citep{moravec1996TR}. 
The main method used to \revf{update} the occupancy values of the voxels was presented by~\citet{thrun2002robotic} and is based on the Log-Odds approach. 
We provide a detailed description of this method in Sec.~\ref{sec:background}. \rev{Another grid-based approach is the histogram grid introduced in~\cite{borenstein1991real} where the cells store certainty values that enable a pseudo-probabilistic representation of occupancy.}

Another class of mapping methods that have shown great success is the class of Gaussian Process-based mapping methods (\eg \cite{OCallaghan2009contextual,senanayake2017learning,OCallaghan2012-IJRR, ramos2016hilbert,kim2014recursive,kim2013occupancy,wang2016fast}). These methods do not rely on voxel grids but model the occupied spaces continuously \rev{given spatial samples of free and occupied areas in the environment}. Furthermore, they take into account the spatial correlations between occupancy of different regions of the map. \citet{schaefer2018dct} present a map representation that stores the map parameters of a decay rate sensor model in the discrete frequency domain.

Different researchers have studied the drawbacks of the Log-Odds approach in occupancy grids and proposed methods to alleviate them \citep{pagac1996evidential, konolige1997improved, Paskin05, veeck2004learning, thrun2003learning, hahnel2003map}. All these methods attempt to mitigate the negative effects caused by the incorrect voxel-independence assumption in mapping. 
In particular, \citet{thrun2003learning} proposes a grid-mapping method using forward sensor models, which takes into account all voxel dependencies and generates maps of higher quality compared to maps resulting from an ISM. However, this method requires the measurement data to be collected offline. Then, it runs the expectation-maximization (EM) algorithm on the full data to compute the most likely map. \citet{hahnel2003map} extend the grid-based mapping methods to dynamic environments using a similar sensor model to the one used in this paper. However, this method assumes accurate measurements (\eg coming from a laser range finder). It also uses EM to compute the map, which limits the result to the most likely values and does not provide any confidence measure on the reported values. In this paper, we propose a mapping method that is online and can cope with high-noise range measurements by incorporating the noise into the model. More importantly, the proposed method computes a confidence value for the estimate which can be very beneficial for planning purposes.

\section{Occupancy grid mapping using inverse sensor models}\label{sec:background}
Most occupancy grid mapping methods decompose the full mapping problem into many binary estimation problems on individual voxels assuming full independence between voxels. This assumption leads to inconsistencies in the resulting map. In this section, we discuss the method and these assumptions.

Let $ G = [G^{1},\cdots,G^{n}] $ be an $ n $-voxel grid overlaid on the \rev{3D (or 2D) environment, where $ G^{i}\in\mathbb{R}^{3} $} is a 3D point representing the center of the $ i $-th voxel of the grid in the global coordinate frame. An occupancy map $ m = [m^{1},\cdots,m^{n}]$ is defined as a set of values over this grid. We start with a more general definition of occupancy where $ m^{i}\in[0,1]$ denotes what proportion of a voxel is occupied. $ m^{i}=1 $ when the $ i $-th voxel is fully occupied and $ m^{i}=0 $ when it is free. %
For maps where occupancy can only be 0 or 1, we use the notation $^{bin}m^i$ for the occupancy of voxel $i$ to explicitly show that the occupancy is binary, \ie $^{bin}m^i\in\{0,1\}$.

The full mapping problem is defined as estimating map $ m $ based on obtained measurements and robot poses. We denote the sensor measurement at the $ k $-th time step by $ z_{k} $ and the sensor configuration at the $ k $-th time step by $ x_{k} $. As we will discuss further in Sec.~\ref{sec:sensorModel}, $x_k$ characterizes the field of view (\eg a pixel cone or a measurement ray when dealing with a single pixel sensor). Formulating the problem in a Bayesian framework, we compress the information obtained from past measurements $ z_{0:k}=\{z_{0},\cdots,z_{k}  \} $ and $ x_{0:k}=\{x_{0},\cdots,x_{k}  \} $ to create a probability distribution (belief) $ \bar{b}^{m}_{k} $ on the map $ m $.
\begin{align}
\bar{b}^{m}_{k}=p(m|z_{0:k},x_{0:k})
\end{align}

However, due to challenges in storing and updating such a high-dimensional belief, grid mapping methods start from individual cells (marginal distributions).
\begin{assumption}
	\pr{Collection of marginals}
	The map pdf is represented by the collection of individual voxel pdfs (marginal pdfs), instead of the full joint pdf.
	\begin{align}
	b^{m}_{k}\equiv(b^{m^{i}}_{k})_{i=1}^{n},~~~~~
	b^{m^{i}}_{k}=p(m^{i}|z_{0:k},x_{0:k})
	\end{align}
	where $ n $ denotes the number of voxels in the map.
	\label{assump:cell_indep}
\end{assumption}

To compute the marginal $ b^{m^{i}} $ in a recursive manner, the method starts \revf{by} applying Bayes' rule.

\begin{align}
\label{eq:firstBayes}
b^{m^{i}}_{k}&=p(m^{i}|z_{0:k},x_{0:k})=
\frac{p(z_{k}|m^{i},z_{0:k-1},x_{0:k})p(m^{i}|z_{0:k-1},x_{0:k})}{p(z_{k}|z_{0:k-1},x_{0:k})}
\end{align}

The main incorrect assumption is applied here:
\begin{assumption}
	\pr{(Incorrect) Measurement independence}
	Standard approaches assume that a single voxel is sufficient (independent of other voxels) to characterize the measurement. Mathematically, 
	\begin{align}
	p(z_{k}|m^{i},z_{0:k-1},x_{0:k})\approx p(z_{k}|m^{i},x_{k})\label{eq:wrong_independence}
	\end{align}
	\label{assump:measurement_independence}
\end{assumption}

\begin{Remark}
	Note that Assumption \ref{assump:measurement_independence} (Eq. \ref{eq:wrong_independence}) would be precise if conditioning was over the whole map. In other words, 
	\begin{align}
	p(z_{k}|m,z_{0:k-1},x_{0:k})= p(z_{k}|m,x_{k})
	\end{align}
	is correct. But, when conditioning on a single voxel, this approximation could be very inaccurate, because a single voxel $ m^{i} $ is not enough to generate the likelihood of observation $ z $. For example, there might even be a wall between $ m^{i} $ and the sensor, and clearly $ m^{i} $ alone cannot tell what range will be measured by the sensor in that case.
\end{Remark}

When dealing with noisy sensors such as stereo cameras, or sensors with large measurement cones such as sonar (even in the absence of any noise), this assumption leads to map conflicts and estimation inconsistencies.

\pr{Inverse sensor model}
Following Assumption~\ref{assump:measurement_independence}, we can apply Bayes' rule to Eq.~\ref{eq:wrong_independence}
\begin{align}\label{eq:inverseSensorModel}
p(z_{k}|m^{i},x_{k})=
\frac{p(m^{i}|z_{k},x_{k})p(z_k|x_k)}{p(m^{i}|x_{k})}
\end{align}
which gives rise to the concept of the \emph{inverse sensor model} (ISM), \ie $ p(m^{i}|z_{k},x_{k}) $. The inverse sensor model describes the occupancy probability given a single measurement. The model cannot be derived from a sensor model. However, depending on the application and the utilized sensor, ad-hoc models can be hand-engineered. The reason to create this model is that it leads to an elegant mapping scheme on binary maps as follows.

\revf{Substituting}~\eqref{eq:wrong_independence} and~\eqref{eq:inverseSensorModel} into~\eqref{eq:firstBayes}, we obtain:
\begin{align}
\nonumber
b^{m^{i}}_{k}&=p(m^{i}|z_{0:k},x_{0:k}) \\
\nonumber
&=\frac{p(m^{i}|z_{k},x_{k})p(z_k|x_k)p(m^{i}|z_{0:k-1},x_{0:k})}{p(m^{i}|x_{k})p(z_{k}|z_{0:k-1},x_{0:k})}
\intertext{Given that the robot's motion does not affect the map:}
\label{eq:blahblah}
b^{m^{i}}_{k} &=
\frac{p(m^{i}|z_{k},x_{k})p(z_k|x_k)p(m^{i}|z_{0:k-1},x_{0:k-1})}{p(m^{i})p(z_{k}|z_{0:k-1},x_{0:k})}
\end{align}

\begin{assumption}
	\pr{Binary occupancy}
	To complete the recursion, traditional grid-based methods further assume that the occupancy of voxels are binary, \ie \mbox{$ ^{bin}m^{i} \in\{0,1\} $.}\\Thus, $ p(^{bin}m^{i}=1)=1-p(^{bin}m^{i}=0) $.
	\label{assump:binary_occupancy}
\end{assumption}

According to Assumption \ref{assump:binary_occupancy}, one can define odds $ r^{i}_{k} $ of occupancy and compute it using Eq.~\eqref{eq:blahblah}:
\begin{align}
\label{eq:logOdds_definition}
r^{i}_{k}&:=\frac{p(^{bin}m^{i}=1|z_{0:k},x_{0:k})}{p(^{bin}m^{i}=0|z_{0:k},x_{0:k})} \\
\nonumber
&=\frac{p(^{bin}m^{i}=1|z_{k},x_{k})p(^{bin}m^{i}=0)}{p(^{bin}m^{i}=0|z_{k},x_{k})p(^{bin}m^{i}=1)}r^{i}_{k-1}
\end{align}

\begin{Remark}
	Relying on Assumption \ref{assump:binary_occupancy} and using odds removes the terms which are difficult to compute from the recursion in Eq.~\eqref{eq:blahblah}. %
\end{Remark}

Further, denoting Log-Odds as $ l_{k}^{i}=\log r^{i}_{k} $, we can simplify the recursion as:
\begin{align}\label{eq:logOdds-recursion}
l^{i}_{k}&=l^{i}_{k-1}+l^{i}_{ISM}-l^{i}_{prior}
\end{align}
where, $l_{ISM}^{i}=\log(p(^{bin}m^{i}=1|z_{k},x_{k})p(^{bin}m^{i}=0|z_{k},x_{k})^{-1})$ are the Log-Odds of the ISM at voxel $ i $, and $ l^{i}_{prior}=\log(p(^{bin}m^{i}=1)p(^{bin}m^{i}=0)^{-1})$ are the Log-Odds of \revf{the} prior. The ISM is often hand-engineered for a given sensor and environment. Fig.~\ref{fig:invSensorModel} shows the typical form of the ISM function.
\ifthenelse{\ifArxiv}{
\begin{figure}[ht!]
	\centering
	\includegraphics[width=0.75\columnwidth]{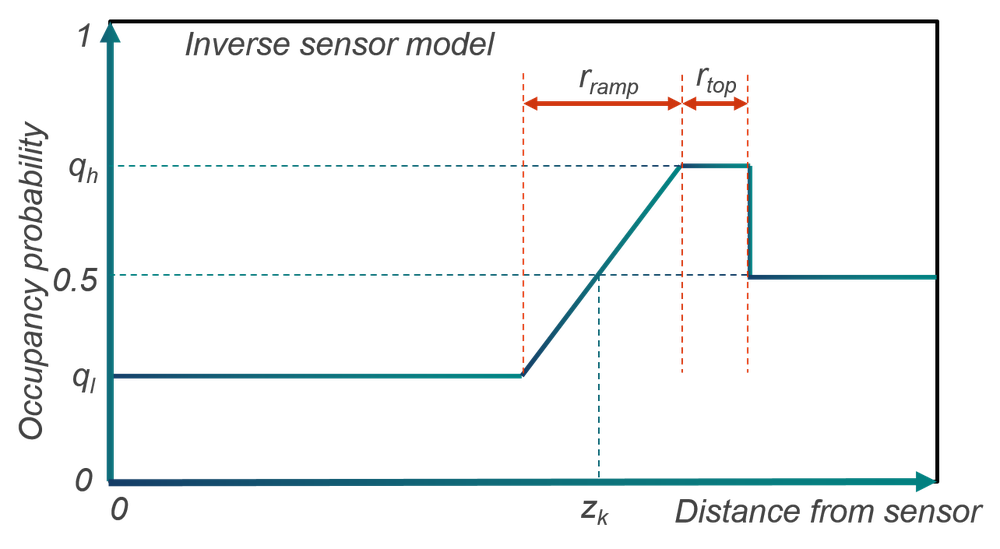}
	\caption{Typical inverse sensor model for a range sensor. It returns the occupancy probability for voxels on the measurement ray/cone based on their distance to the camera.}
	\label{fig:invSensorModel}
\end{figure}
}{} %

\section{Confidence-rich Representation} \label{subsec:problem}
In our map representation we store the probability distribution of $ m^{i} $ in each voxel $ i $. The variable $ m^{i} $ in this paper can be interpreted in two ways:
\begin{enumerate}[leftmargin=0cm,itemindent=.5cm,labelwidth=0.4cm,labelsep=0cm,align=left]
		\item In the more general setting, $ m^{i}\in[0,1] $ directly represents the occupancy level (the fraction of voxel $ i $ that is occupied by obstacles). The proposed method can model continuous occupancy and Assumption~\ref{assump:binary_occupancy} in traditional occupancy mapping can be relaxed.
		\item If the underlying true map is assumed to be a binary map (denoted by $^{bin}m$), the occupancy of the $ i $-th voxel $ ^{bin}m^{i}\in\{0,1\} $ follows the Bernoulli distribution $ ^{bin}m^{i}\sim \operatorname{Bernoulli}(m^{i}) $. In other words, in this case $ m^{i} $ refers to the parameter of the Bernoulli distribution. While ISM-based mapping methods store $ m^{i} $ as a deterministic value, we estimate $ m^{i} $ probabilistically based on measurements and store its pdf at each voxel. Note that in this setting, $ ^{bin}m^{i}\in\{0,1\} $, where $ m^{i}\in[0,1] $ represents the occupancy probability, \ie $ m^{i}_{k}=p(^{bin}m^{i}=1|z_{0:k},x_{0:k}) $.
\end{enumerate}

\noindent
\pr{Problem description}
Given the above-mentioned representation, we aim at estimating $ m $ based on noisy measurements by computing its posterior distribution. We define three beliefs over the map: (1) full map belief $ \bar{b}^{m}_{k}=p(m|z_{0:k},x_{0:k}) $, (2) marginal cell beliefs $ b^{m^{i}}_{k}=p(m^{i}|z_{0:k},x_{0:k})$, and (3) the collection of marginals $b^{m}_{k}=(b^{m^{i}}_{k})_{i=1}^{M}$\footnote{More precisely, in these definitions, the variable $b$ refers to the set of parameters that characterize the probability distributions. Hence, we will treat $b$ as a vector (deterministic or random depending on the context) in the rest of the paper.}.
Similar to ISM-based methods, for mapping, we maintain and update the collection of marginals $ b^{m}_{k} $. To do so, we  derive the following items:
\begin{enumerate}[leftmargin=0cm,itemindent=.5cm,labelwidth=0.4cm,labelsep=0cm,align=left]
	\item \pr{Ranging sensor model}
	Given that the obstacles are described by a stochastic map, we derive a ranging sensor model, \ie the probability of obtaining measurement $ z $ given a stochastic map and robot location: $ p(z_{k}| x_{k}, b^{m}_{k}) $. This model will be used in the map update module.
	
	\item \pr{Recursive density mapping}
	We derive a recursive mapping scheme $ \tau $ that updates the current density map based on the last measurements
	\begin{align}\label{eq:recursiveMapping}
	b^{m^{i}}_{k+1}=\tau^{m^{i}}(b^{m}_{k},z_{k+1},x_{k+1}).
	\end{align}
	
	The fundamental difference with ISM-mapping is that the evolution of the $ i $-th voxel depends on other voxels as well. Note that the input argument to $ \tau^{m^{i}} $ is the collection of all voxel beliefs $ b^{m} $, not just the $ i $-th voxel belief $ b^{m^{i}} $. 

\end{enumerate}

While motion planning is beyond the scope of this paper, we briefly discuss how planning can benefit from this enriched map data and consistent estimation mechanism to 1) predict the collision risk more accurately and 2) generate actions that actively reduce mapping uncertainty and lead to safer paths.

\rev{\pr{Remark} In relation to the traditional occupancy methods (see Section~\ref{sec:background}), the proposed method relaxes Assumptions~\ref{assump:measurement_independence} and \ref{assump:binary_occupancy} of the ISM-based mapping.}

\section{Range-sensor Modeling} \label{sec:sensorModel}
In this section, we model a range sensor when the environment representation is a stochastic map. We focus on passive ranging sensors like stereo cameras, but the discussion easily translates to active sensors as well. 

\pr{Ranging pixel}
Let us consider an array of ranging sensors (\eg disparity pixels). We denote the camera center by $ x_{cam} $, the 3D location of the $ i $-th pixel by $ v $, and the ray emanating from $ x_{cam} $ and passing through $ v $ by $ x = (x_{cam},v) $. Let $ r $ denote the distance between the camera center and the closest obstacle to the camera along ray $ x $. 

\rev{{\it Example:} For a stereo camera, the range $ r $ is related to the measured disparity $ z $ as $z = r^{-1} f d_b$~\cite{Hartley2000}, where $ f $ is the camera's focal length and $ d_b $ is the baseline between two cameras on the stereo rig.}

In the following, we focus on a single pixel $ v $ and derive the forward sensor model $ p(z|x,b^{m}) $.

\begin{figure}[ht!]
	\centering
	\includegraphics[width=0.75\textwidth]{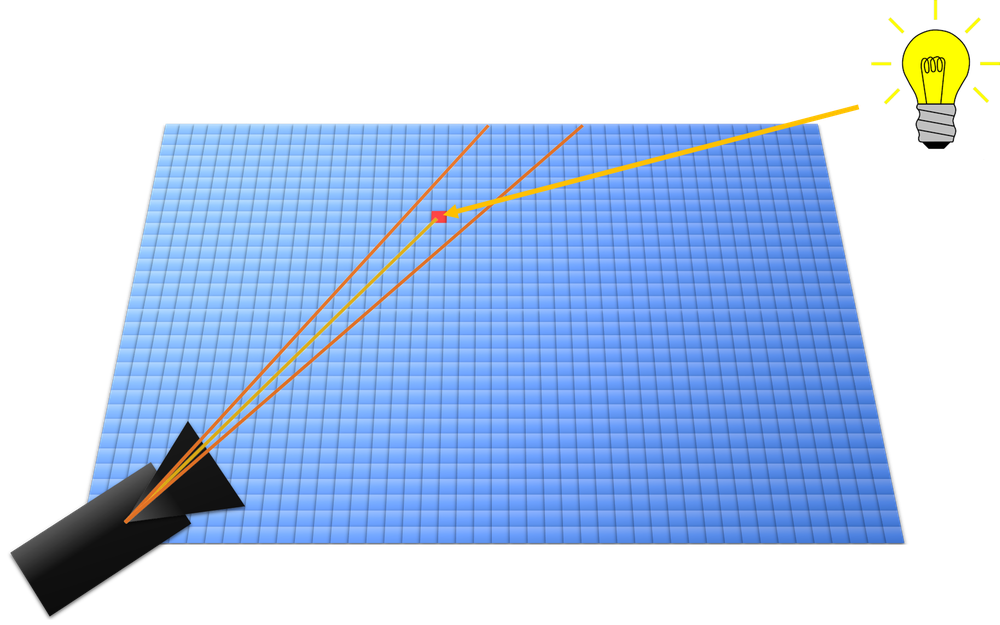}
	\caption{The cone formed by two red lines depicts the field of view of pixel $ v $. The disparity measurement on pixel $ v $ can be caused by light bouncing off any of the voxels in the pixel cone and reaching the image plane. In this figure, the measurement is created by light bouncing off the voxel highlighted in red.}
	\label{fig:grid}
\end{figure}

\begin{figure}[ht!]
	\color{blue}
	\centering
	\includegraphics[width=0.85\textwidth]{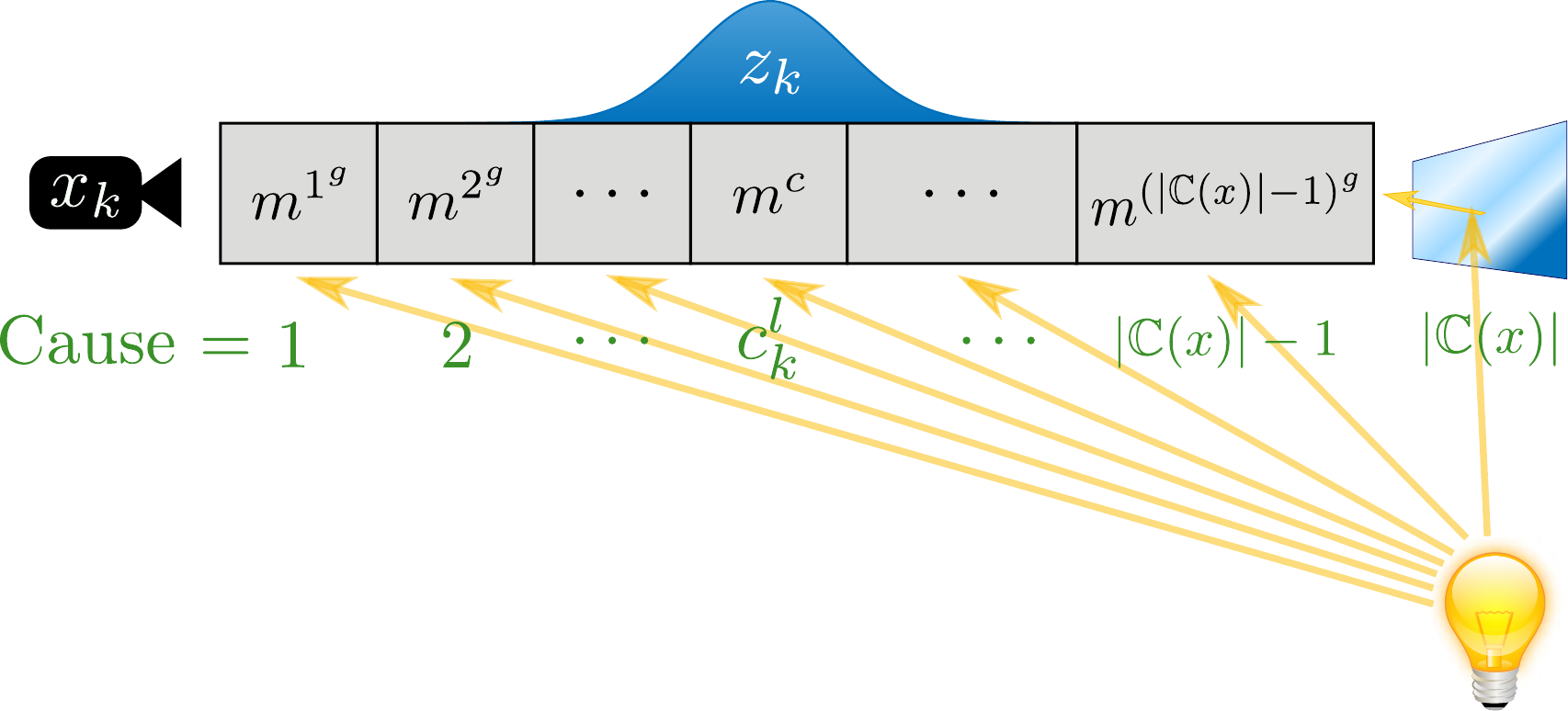}
	\caption{\rev{Illustration of the Sensor Cause Model (SCM). $x_k$ is the sensor position and the figure shows a ray of voxels in the direction of the sensor. 
	In this simple model, light (yellow arrows) reaches the sensor directly, or by bouncing off of any of voxels on the ray. For unification, the latter case is modeled as bouncing off of an abstract voxel number $|\mathbb{C}(x)|$, which lies outside the map boundaries. Given a cause voxel (e.g., $c$), the forward sensor model can be characterized easily (here visualized by a blue Gaussian distribution).}}
	\label{fig:scm}
\end{figure}

\pr{Pixel cone}
Consider the field of view of pixel $ v $. More precisely, it is a narrow 3D cone with apex at $ x $ and boundaries defined by pixel $ v $ \rev{(see Fig. \ref{fig:grid})}. Also, for simplicity, one can consider just a ray $ x $ going through camera center $ x $ and the center of pixel $ v $ \rev{(see Fig.~\ref{fig:scm})}. The pixel cone $ \mathtt{Cone}(x) $ refers to the set of voxels in map $ m $ that \revf{intersect} this cone (or intersect with ray $ x $). We denote this set by $ \mathbb{C}^{vox}=\mathtt{Cone}(x) $. \rev{For the reasons that will be clear below, we augment the cone with one abstract voxel, referred to as $c^{light}$. This voxel will capture the case where light reaches the camera directly from the light source. Accordingly, we denote the augmented cone as $ \mathbb{C}(x)=(\mathbb{C}^{vox},c^{light}) $.}

\pr{Local vs. global indices}
For a given ray $ x $, we order the voxels along the ray from closest to farthest from the camera. Notation-wise, $ i^{l}\in\{1,\cdots,|\mathbb{C}|\} $ represents the local index of a voxel on ray $ x $. Function $ i^{g} = g(i^{l},x) $ returns the global index $ i^{g} $ of this voxel in the map.

\pr{Cause variables}
The disparity measurement on pixel $ v $ could be the result of light bouncing off any of the voxels in the cone \revf{$ \mathbb{C}^{vox}=\mathtt{Cone}(x) $} (see Fig.~\ref{fig:grid} and \ref{fig:scm}). Therefore, any of these voxels are a potential cause for a given measurement. In the case that the environment map is perfectly known, one can pinpoint the exact cause by finding the closest obstacle to the camera center. However, when the knowledge about the environment is partial and probabilistic, the best \revf{that can be deduced} about causes is a probability distribution over all possible causes in the \revf{augmented pixel cone $ \mathbb{C}(x) $}. These causes will play an important role (as hidden variables) in deriving the sensor model for stochastic maps.

\pr{Cause probability}
To derive the full sensor model, we need to reason about which voxel was the cause for a given measurement. For a voxel $ c\in\mathbb{C}(x) $ to be the cause, two events need to happen: \textit{(i)} $ B^c $, which indicates the event of light bouncing off voxel $ c $ and \textit{(ii)} $ R^c $, which indicates the event of light reaching the camera from voxel $ c $.
\begin{align}
p(c | b^{m}) = \Pr(B^{c}\!,R^{c}| b^{m}) 
=
\Pr(R^{c}|B^{c}\!, b^{m})\!\Pr(B^{c}| b^{m})
\end{align}

\pr{Bouncing probability}
To compute the bouncing probability, we rely on the fact that $ \Pr(B^{c}|m^{c}) = m^{c} $ (by definition). %
Note that $ \Pr(B^{c}|\,m^{c}\!,\,b^{m}) = \Pr(B^{c}|m^{c}) $.
\begin{align}
\Pr(B^{c}| b^{m}) &= \int_{0}^{1}\Pr(B^{c}|\,m^{c}\!,\,b^{m})
p(m^{c}| b^{m}) ~\text{d}m^{c} \\
\nonumber
&=\int_{0}^{1}m^{c}
b^{m^{c}} ~\text{d}m^{c}
=\mathbb{E}{m^{c}}=\widehat{m}^{c}
\end{align}

\pr{Reaching probability}
For the ray emanating from voxel $ c $ to reach the image plane, it has to go through all voxels on ray $ x $ between $ c $ and the sensor. Let $ c^{l} $ denote the local index of voxel $ c $ along the ray $ x $, \ie $ c^{l}=g^{-1}(c,x) $. Then we have:
\begin{align}
\Pr&(R^{c}| \,B^{c}\!,\, b^{m}) \\
&= 
\nonumber
(1 - \Pr(B^{g(c^{l}- 1,x)}|\, b^{m}))
\Pr(R^{g(c^{l}- 1,x)}|\, B^{g(c^{l}- 1,x)}\!, b^{m})
\\
\nonumber
&= \prod_{l = 1}^{c^{l}-1}
(1 - \Pr(B^{g(l,x)}|\, b^{m}))
= \prod_{l = 1}^{c^{l}-1}
(1 - \widehat{m}^{g(l,x)})
\end{align}

\pr{Sensor model with known cause}
\rev{Assuming the cause voxel for measurement $ z $ is known, the forward sensor model $z = h(x,c,n_{z})$ is typically a function of the distance between the cause voxel and the sensor $\|G^{c}-x_{cam}\|$ contaminated by measurement noise $n_{z}$. It can also be \revf{represented by} a likelihood function $p(z|x,c)$.}

\rev{{\it Example:} For a stereo camera, the sensor model with known cause can be simplified as:
\begin{align}
z = h(x,c,n_{z}) = \|G^{c}-x_{cam}\|^{-1}fd_{b}+n_{z},
\end{align}
where, $ n_{z}\sim\mathcal{N}(0,V) $ denotes the observation noise, drawn from a zero-mean Gaussian with variance $ V $. We can alternatively describe the observation model in terms of pdfs as follows:
\begin{align}
p(z | x,c) = \mathcal{N}(\|G^{c}-x_{cam}\|^{-1}fd_{b}, V).
\end{align}
}

\pr{Sensor model with stochastic maps}
The sensor model given a stochastic map can be computed by incorporating hidden cause variables $\mathbb{C}(x)$ into the formulation:
\begin{align}
p&(z | x, b^{m}) \\
\nonumber
&=
\sum_{c\in\mathbb{C}(x)}p(z|x,c,b^{m})p(c|b^{m})
=
\sum_{c\in\mathbb{C}(x)}p(z|x,c)p(c|b^{m})
\\
\nonumber
&=
\sum_{c\in\mathbb{C}(x)}
p(z|x,c)
\widehat{m}^{c}
\prod_{l = 1}^{c^{l}-1}
(1 - \widehat{m}^{g(l,x)})
\end{align}

\rev{{\it Example:} 
In the simplified stereo camera case, the sensor model can be written as:
\begin{align}
p(z | x, b^{m})
\nonumber
=
\!\!\!
\sum_{c\in\mathbb{C}(x)}
\!\!
\mathcal{N}(\|G^{c}-x_{cam}\|^{-1}fd_{b}, V)
\widehat{m}^{c}
\prod_{l = 1}^{c^{l}-1}
(1 - \widehat{m}^{g(l,x)})
\end{align}
}

\section{Confidence-Augmented Grid Map} \label{sec:mapping}
In this section, we derive the recursive mapping algorithm described in Eq.~\eqref{eq:recursiveMapping}. This mapping algorithm $\tau$ can not only reason about the occupancy at each cell but also about the confidence level of this value. As a result, it enables efficient prediction of the map that can be embedded in motion planning, resulting in safer trajectories. 
\ifthenelse{\ifArxiv}{We start with a lemma that will be used in the following derivations. %
\begin{Lem}\label{lem:cause-meas}
	Given the cause, the value of the corresponding measurement is irrelevant.
	\begin{align}
	\nonumber
	p(m^i|c_k,z_{0:k},x_{0:k})
	=
	p(m^i|c_{k},z_{0:k-1},x_{0:k}).
	\end{align}
\end{Lem}
}{} %
\revf{
\begin{proof}
The proof follows from Bayes' rule and the fact that cause $c_k$ is a sufficient statistic for observation $z_k$, i.e., $p(z_k|c_k,m^i,z_{0:k-1},x_{0:k}) = p(z_k|c_k)$.\qed
\end{proof}
}

To compute the belief of the $ i $-th voxel, denoted by $ b^{m^{i}}_{k} = p(m^i|z_{0:k},x_{0:k}) $, we bring the cause variables into the formulation.
\begin{align}
\label{eq:BayesianUpdate}
b^{m^{i}}_{k} 
&= p(m^i|z_{0:k},x_{0:k}) \\
\nonumber
&=
\!\!\!\!\!
\sum_{c_{k}\in\mathbb{C}(x)}
\!\!\!\!\!
p(m^i|c_k,z_{0:k},x_{0:k})p(c_k|z_{0:k},x_{0:k})
\\
\nonumber
&=
\!\!\!\!\!
\sum_{c_{k}\in\mathbb{C}(x)}
\!\!\!\!\!
p(m^i|c_k,z_{0:k-1},x_{0:k})p(c_k|z_{0:k},x_{0:k})
\\
\nonumber
&=
\!\!\!\!\!
\sum_{c_{k}\in\mathbb{C}(x)}
\!\!\!\!\!
\frac{
	p(c_k|m^i,z_{0:k-1},x_{0:k})
}
{p(c_k|z_{0:k-1},x_{0:k})}
p(c_k|z_{0:k},x_{0:k})
b^{m^{i}}_{k-1}
\end{align}

It can be shown that $ b_{k-1}^{m} $ is a sufficient statistic \citep{kay1993fundamentals} for the data $ (z_{0:k-1},x_{0:k-1}) $ in the above terms. \revf{Intuitively, this means $ b_{k-1}^{m} $ captures all the available information about $c_k$ from the data $ (z_{0:k-1},x_{0:k-1}) $.
} Thus, we can rewrite \eqref{eq:BayesianUpdate} as:
\begin{align}
b^{m^{i}}_{k}
\!\!
=
\!\!\!\!\!\!\!
\sum_{c_{k}\in\mathbb{C}(x)}
\!\!\!\!\!\!
\frac
{p(c_k|m^i,b^{m}_{k-1},x_{k})}
{p(c_k|b^{m}_{k-1},x_{k})}
\!
p(c_k|b^m_{k-1},z_{k},x_{k})
b^{m^{i}}_{k-1}
\label{eq:recur-with-ratio}
\end{align}

In the following, we make the assumption that the map pdf is sufficient for computing the bouncing probability from voxel $ c $ (\ie one can ignore voxel $ i $ given the rest of the map).
Mathematically, for $ c_{k}\neq i $ we assume:
\begin{align}
\nonumber
&\Pr(B^{c_{k}}| m^{i}, b^{m}_{k-1},x_{k})
\approxeq \Pr(B^{c_{k}}| b^{m}_{k-1},x_{k})
= \widehat{m}^{c_{k}}
\end{align}

\begin{figure}[ht!]
    \centering
    \includegraphics[width=.9\columnwidth]{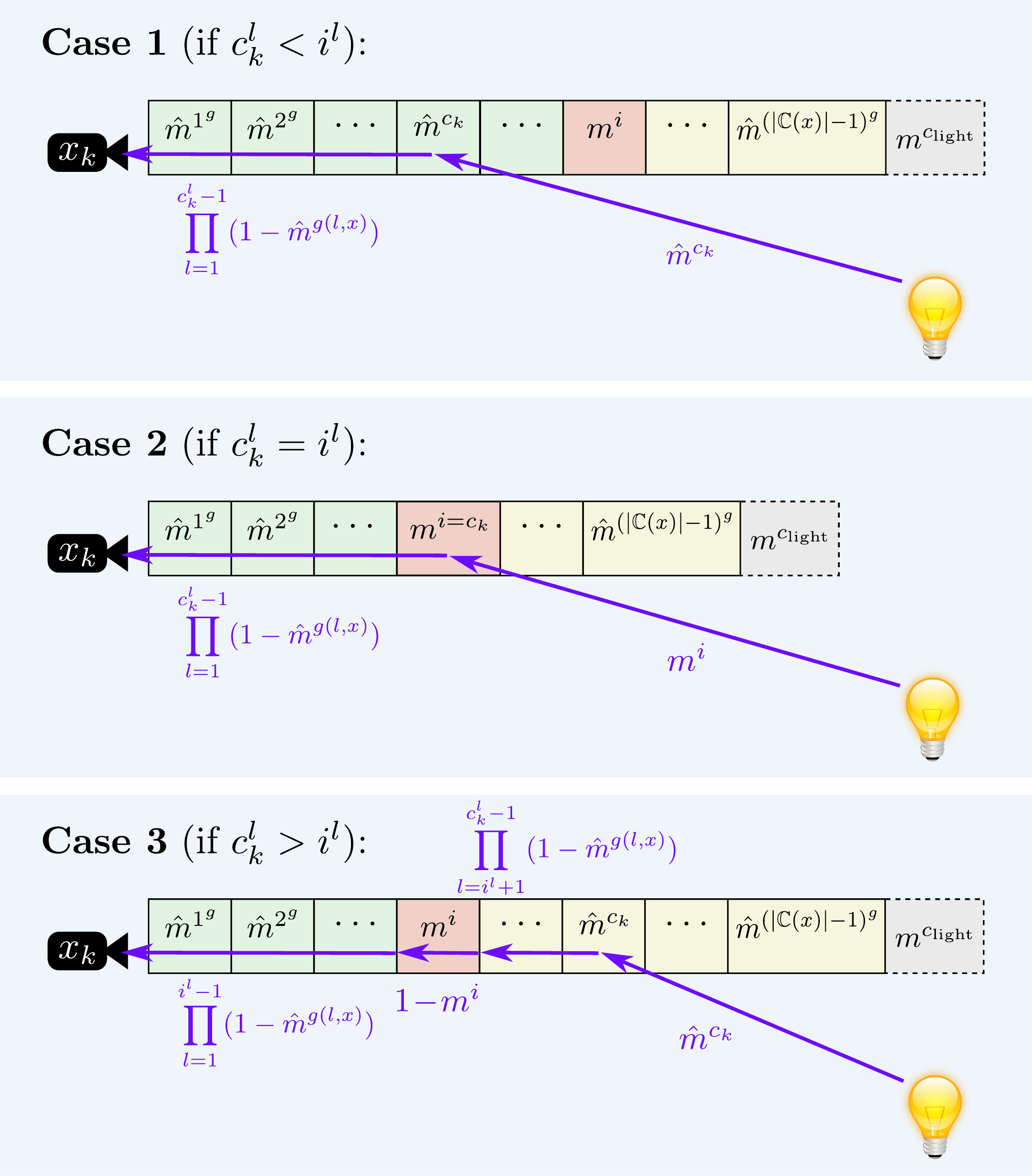}
    \caption{\rev{Visualization of the three cases in Eq.~\eqref{eq:three-piece} that models the sensor cause model given full knowledge about $m^i$ and belief about the rest of voxels. %
    We follow the convention in Fig.~\ref{fig:scm}. Purple arrows depict the light rays. Here, green and yellow voxels represent the voxels that are, respectively, closer and farther to the camera than the voxel $m^i$. In cases 1 and 2, we have two terms corresponding to bouncing and reaching probabilities. In case 3, we have four terms in Eq.~\eqref{eq:three-piece} that correspond to the bouncing probability and the probability of passing through yellow, red, and green voxels.
    }}
    \label{fig:three-piece}
\end{figure}

Note that we still preserve a strong interdependence between voxels via the reaching probability. To see this clearly, we expand the numerator $ p(c_{k} | m^{i}, b^{m}_{k-1},x_{k}) $ in Eq.~\eqref{eq:recur-with-ratio} (we drop $ x $ to unclutter the equations) as:
\begin{align}
\label{eq:three-piece}
p&(c_{k} | m^{i}, b^{m}_{k-1}) \\
\nonumber
&=
\Pr(B^{c_{k}},R^{c_{k}}| m^{i}, b^{m}_{k-1})\\
\nonumber
&=
\Pr(B^{c_{k}}| m^{i}, b^{m}_{k-1})\Pr(R^{c_{k}}|B^{c_{k}}, m^{i}, b^{m}_{k-1})\\
\nonumber
&= \begin{cases}
\widehat{m}^{c_{k}}
\prod_{l = 1}^{c^{l}_{k}-1}
(1 - \widehat{m}^{g(l,x)})
& \text{if       } c^{l}_k < i^{l}\\[6pt]
m^{i}
\prod_{l = 1}^{c^{l}_{k}-1}
(1 - \widehat{m}^{g(l,x)})
& \text{if       } c^{l}_k = i^{l}\\[6pt]
\parbox[]{5cm}
{
	$
	\widehat{m}^{c_{k}}
	\left(\prod_{l = 1}^{i^{l}-1}
	(1 - \widehat{m}^{g(l,x)})\right)\\
	~~~\times(1-m^{i})
	\left(\prod_{l = i^{l}+1}^{c^{l}_{k}-1}
	(1 - \widehat{m}^{g(l,x)})\right)
	$
}
& \text{if       } c^{l}_k > i^{l}
\end{cases}
\end{align}

\rev{The first line in Eq.~\eqref{eq:three-piece} captures the probability of the cause voxel $c_k$ when voxel $i$ is not on the the light trace (see Fig.~\ref{fig:three-piece}). Thus, it is composed of \revf{the} probability of bouncing back from the cause voxel $\widehat{m}^{c_{k}}$ and passing through all voxels on the ray to reach the sensor $\prod_{l = 1}^{c^{l}_{k}-1}
(1 - \widehat{m}^{g(l,x)})$. The second line in Eq.~\eqref{eq:three-piece} captures the case where the $i$-th voxel is the cause voxel and the third line in Eq.~\eqref{eq:three-piece} captures the last case where the $i$-th voxel is between the cause voxel and the sensor. In this case, the probability of reaching the sensor from the cause voxel will explicitly use the occupancy information of the voxel $i$.}

\rev{The denominator in Eq.~\eqref{eq:recur-with-ratio} is $ p(c_{k} | b^{m}_{k-1},x_{k}) = \widehat{m}^{c_{k}}
\prod_{l = 1}^{c^{l}_{k}-1}
(1 - \widehat{m}^{g(l,x)})
$ for all $ c_{k}\in\mathbb{C}(x) $.} In these equations, $ c^{l}_{k}=g^{-1}(c_{k},x_{k}) $ and $ i^{l}=g^{-1}(i,x_{k}) $ are the corresponding indices of $ c_{k} $ and $ i $ in the local frame.

Therefore, the ratio in \eqref{eq:recur-with-ratio} is simplified to:
\ifthenelse{\ifISRR}{
\begin{align}
\nonumber
\frac
{p(c_k|m^i,b^{m}_{k-1},x_{k})}
{p(c_k|b^{m}_{k-1},x_{k})}
= \mathbb{I}(c^{l}_k < i^{l})+
m^{i}(\widehat{m}^{i})^{-1}
\mathbb{I}(c^{l}_k = i^{l})
+
(1-m^{i})(1-\widehat{m}^{i})^{-1}
\mathbb{I}(c^{l}_k > i^{l})
\end{align}
where $\mathbb{I}(A)$ equals one if event $A$ is true and zero otherwise.
} %
{
\begin{align}
\nonumber
\frac
{p(c_k|m^i,b^{m}_{k-1},x_{k})}
{p(c_k|b^{m}_{k-1},x_{k})}
&= \begin{cases}
1
&\text{if } c^{l}_k < i^{l}\\
m^{i}(\widehat{m}^{i})^{-1}
&\text{if } c^{l}_k = i^{l}\\
(1-m^{i})(1-\widehat{m}^{i})^{-1}
&\text{if } c^{l}_k > i^{l}\\
\end{cases}
\end{align}
} %
Plugging the ratio back into \eqref{eq:recur-with-ratio} and collecting linear and constant terms, we can show that:
\begin{align}
&p(m^i|z_{0:k},x_{0:k})=
(\alpha^{i} m^{i}+\beta^{i})p(m^i|z_{0:k-1},x_{0:k-1})
\label{eq:final-mapping-alphaBeta}
\end{align}

\noindent where
\begin{align}\label{eq:alpha}
\alpha^{i} &=(\widehat{m}^{i})^{-1}
p(c_k|b^m_{k-1},z_{k},x_{k}) \\
\nonumber
&~~~~
-
(1-\widehat{m}^{i})^{-1}
\sum_{c^{l}_{k}=i^{l}+1}^{|\mathbb{C}(x)|}
p(c_k|b^m_{k-1},z_{k},x_{k})
\\
\beta^{i} &=
\sum_{c^{l}_{k}=1}^{i^{l}-1}
p(c_k|b^m_{k-1},z_{k},x_{k}) \\
\nonumber
&~~~~
+
(1-\widehat{m}^{i})^{-1}
\sum_{c^{l}_{k}=i^{l}+1}^{|\mathbb{C}(x)|}
p(c_k|b^m_{k-1},z_{k},x_{k})\label{eq:beta}
\end{align}

\noindent
In a more compact form we can rewrite Eq.~\eqref{eq:final-mapping-alphaBeta} as:
\begin{align}
b_{k+1}^{m^{i}}=\tau^{i}(b_{k}^{m},z_{k+1},x_{k+1}).
\end{align}

\noindent
\pr{Sensor cause model}
The proposed machinery gives rise to the term $ p(c_k|z_{0:k},x_{0:k})
=p(c_k|b^m_{k-1},z_k,x_{k}) $, which is referred to as \emph{Sensor Cause Model} (SCM) in this paper. As opposed to the inverse sensor model in traditional mapping that needs to be hand-engineered, the SCM can be derived from the forward sensor model in a principled way as follows (\emph{cf.} Fig.~\ref{fig:scm}). %
\begin{align}\label{eq:sensorCauseModel}
p(c_k|z_{0:k},x_{0:k})
&=p(c_k|b^m_{k-1},z_k,x_{k}),~~~~\forall c_k\in\mathbb{C}(x_k) \\
\nonumber
&=\frac{p(z_k|c_k,x_k)p(c_k|b^m_{k-1},x_{k})}
{p(z_k|b^m_{k-1},x_k)} \\
\nonumber
&=\eta' p(z_k|c_k,x_k)p(c_k|b^m_{k-1},x_{k}) \\
\nonumber
&=\eta' p(z_k|c_k,x_k)\widehat{m}^{c_k}_{k-1}
\!\!\prod_{j=1}^{c^l_k-1}\!\!(1-\widehat{m}^{g(j,x)}_{k-1})
\intertext{\rev{where $$ \eta'=\sum_{c_k\in\mathbb{C}(x)}p(z_k|c_k,x_k)\widehat{m}^{c_k}_{k-1}
\!\!\prod_{j=1}^{c^l_k-1}\!\!(1-\widehat{m}^{g(j,x)}_{k-1}) $$ is the normalization constant.}}
\nonumber
\end{align}
\subsection{Algorithm and Computational Complexity} \label{sec:algorithm}
The complete belief-update algorithm is presented in Alg.~\ref{alg:mapping}. 
For every measurement obtained from a single ray, we find all the voxels $\mathbf{V_{ray}}$ that intersect the ray (line 1) using Bresenham's line algorithm. 
In the next step, the sensor cause model is computed (line 2). 
It is worth noting that, if implemented recursively, this step has linear complexity w.r.t. the number of voxels on the ray, \ie $O(\mathbf{|V_{ray}|})$.
The terms in Eqs.~\eqref{eq:alpha} and~\eqref{eq:beta} have been computed in Line 2 and stored as SCM. In lines 4 and 5, we retrieve these values and compute the likelihood parameters $\alpha,~\beta$, and concurrently update the voxel belief. \revf{The voxel belief is represented using a constant number of weighted particles.}
Doing these steps at the same time enables us to do recursion again, which results in a linear complexity w.r.t. the number of voxels on the ray, \ie $O(\mathbf{|V_{ray}|})$.

\revf{We assume the map is stored in a voxel grid of a fixed size in which we can retrieve the voxel of interest in constant time.} 
The resulting complexity of the proposed algorithm is \revf{$O(\mathbf{|V_{ray}|}^2)$}.
Assuming the same data structure for storing the map, the standard Log-Odds approach results in the complexity $O(\mathbf{|V_{ray}|})$.

It is important to note that the increase in complexity is minimal for the proposed approach, \rev{which leads to similar computation times in comparison to Log-Odds, as we report in Sec.~\ref{sec:results}}.
Therefore, we enrich the information stored in the map without substantially increasing the complexity of the mapping algorithm.

\begin{figure}
    \centering
    \includegraphics[width=0.9\columnwidth]{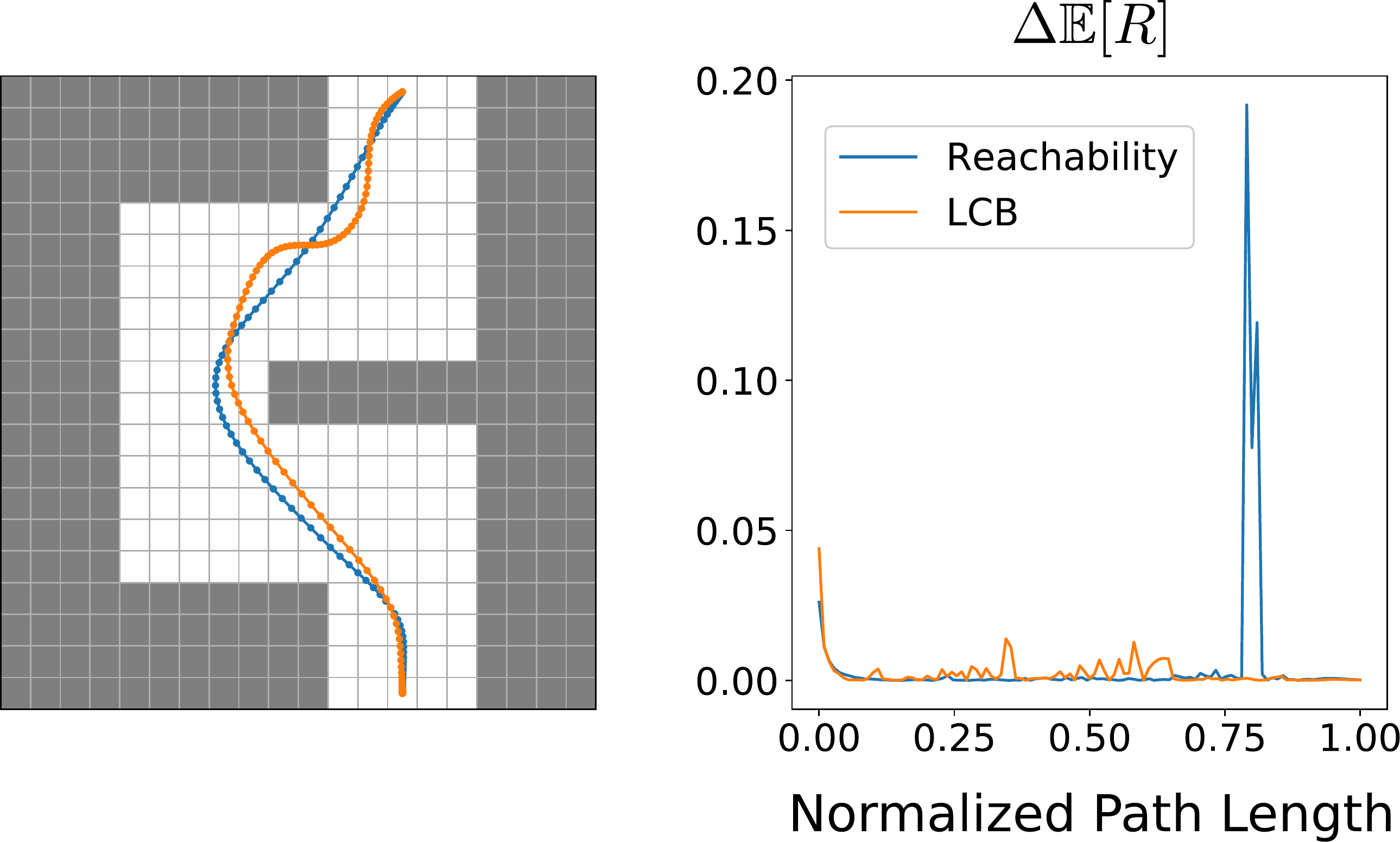}
    \caption{\rev{
    Using the \emph{Lower Confidence Bound} (LCB) as trajectory optimization criterion~\citep{heiden2017planning}, which combines the estimated mean and variance of occupancy in the map, leads to considerably safer trajectories (orange, \emph{left}) with a lower surprise $\Delta\mathbb{E}[R]$ (\emph{right}) in expected reachability throughout the trajectory than trajectories that are optimized purely for expected reachability $\mathbb{E}[R]$ (blue). On the left, the ground-truth map is visualized.}}
    \label{fig:planning}
\end{figure}

\let\oldnl\nl%
\newcommand{\nonl}{\renewcommand{\nl}{\let\nl\oldnl}}%

\begin{algorithm}
	\caption{Confidence-rich grid mapping (CRM)}
	\label{alg:mapping}
	\nonl \textbf{Input:}     Current map belief $ b_{k} $, observation $ z_k $, measurement ray $ x_k $\\
	\nonl \textbf{Output:}    Updated map belief $ b_{k+1} $\\
	\nonl \textbf{Procedure:} $ b_{k+1}=\operatorname{\textsc{Update}}(b_{k},z_{k},x_{k}) $\\
	{
		Find $\mathbf{V_{ray}}$ voxels on the ray $x_k$\\
		Compute SCM using Eq.~\eqref{eq:sensorCauseModel}\\    
		\ForEach{$v_i \in \mathbf{V_{ray}}$}
		{
			Compute $ \alpha^i $, $ \beta^i $ (Eqs. \eqref{eq:alpha}, \eqref{eq:beta})\\
			Update voxel belief (Eq.~\eqref{eq:final-mapping-alphaBeta})\\
		}
		\Return $ b_{k+1} $
	}	
\end{algorithm}

\section{Map Confidence and Safe Exploration for Motion Planning} \label{sec:planning}

\newfloatcommand{capbtabbox}{table}[][11.4cm]
\begin{figure*}[t!]
\begin{floatrow}
\ffigbox[5.8cm]{%
	\includegraphics[height=5cm, trim=10pt 30pt 30pt 30pt]{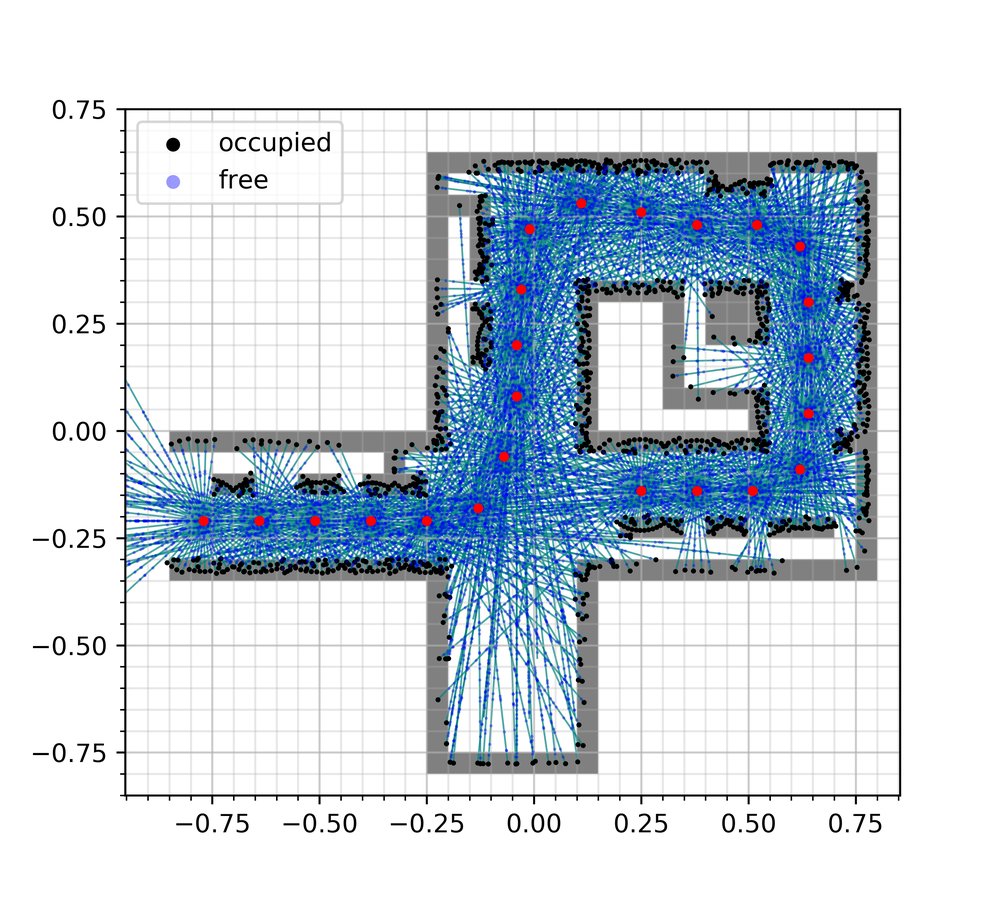}	
}{%
  \caption{\rev{Ground-truth map with sampling sensor locations (red) and samples of free (blue) and occupied (black) space on measurement rays with $0.25s$ sensor noise std. dev. for the experiment described in Sec.~\ref{sec:corridor}. $s$ denotes the voxel size. Units are in meters.}}%
  \label{fig:ground-truth_map}
}\hfill
\capbtabbox{%
\smaller
\rev{
    \begin{tabular}{ l | r r r r r }
\toprule
Sensor std. dev. & \multicolumn{1}{c}{$0.25 s$} & \multicolumn{1}{c}{$0.5 s$} & \multicolumn{1}{c}{$1 s$} & \multicolumn{1}{c}{$2 s$} & \multicolumn{1}{c}{$3 s$} \\
\midrule
	CRM MAE 
		 & \bf 0.368
		 & \bf 0.379
		 & \bf 0.399
		 & \bf 0.416
		 & 0.425
\\
	Log-Odds MAE 
		 & 0.398
		 & 0.401
		 & 0.408
		 & 0.417
		 & 0.419
\\
	Hilbert MAE 
		 & 0.428
		 & 0.440
		 & 0.456
		 & 0.451
		 & 0.463
\\
	GPOM MAE 
		 & 0.532
		 & 0.503
		 & 0.511
		 & 0.448
		 & \bf 0.359
\\[.4em]
    CRM AUC 
		 & \bf 0.970
		 & \bf 0.967
		 & \bf 0.942
		 & \bf 0.793
		 & \bf 0.687
\\
	Log-Odds AUC 
		 & 0.879
		 & 0.842
		 & 0.734
		 & 0.495
		 & 0.379
\\
	Hilbert AUC 
		 & 0.802
		 & 0.813
		 & 0.721
		 & 0.601
		 & 0.560
\\
	GPOM AUC 
		 & 0.460
		 & 0.488
		 & 0.404
		 & 0.336
		 & 0.301
\\[.4em]
	CRM $I_c$ 
		 & \bf 15.914
		 & \bf 18.267
		 & \bf 24.935
		 & \bf 39.303
		 & \bf 46.444
\\
	Log-Odds $I_c$ 
		 & 24.072
		 & 27.350
		 & 35.969
		 & 56.948
		 & 70.332
\\
	Hilbert $I_c$ 
		 & 26.607
		 & 25.580
		 & 33.053
		 & 47.511
		 & 49.629
\\
	GPOM $I_c$ 
		 & 59.639
		 & 59.286
		 & 82.107
		 & 114.357
		 & 133.048
\\[.4em]
	CRM PCC 
		 & \bf 0.984
		 & \bf 0.965
		 & \bf 0.956
		 & \bf 0.968
		 & \bf 0.969
\\
	Log-Odds PCC 
		 & 0.854
		 & 0.874
		 & 0.907
		 & 0.892
		 & 0.855
\\
	Hilbert PCC 
		 & 0.586
		 & 0.614
		 & 0.572
		 & 0.622
		 & 0.504
\\
	GPOM PCC 
		 & -0.028
		 & 0.094
		 & 0.089
		 & 0.512
		 & 0.446
\\[.4em]
	CRM $t$ $[s]$ 
		 & 0.018
		 & 0.018
		 & 0.018
		 & 0.018
		 & 0.018
\\
	Log-Odds $t$ $[s]$ 
		 & \bf 0.015
		 & \bf 0.015
		 & \bf 0.015
		 & \bf 0.015
		 & \bf 0.015
\\
	Hilbert $t$ $[s]$ 
		 & 14.216
		 & 15.501
		 & 16.369
		 & 14.790
		 & 15.566
\\
	GPOM $t$ $[s]$ 
		 & 859.819
		 & 945.756
		 & 1056.777
		 & 1305.687
		 & 1546.306
\\
	\bottomrule
    \end{tabular}
}
}{%
  \caption{\rev{Mean Absolute Error (MAE), inconsistency $I_c$ with $\gamma=\nicefrac{1}{2}$ (Eq.~\ref{eq:inconsistencyMeasure}), Pearson correlation between estimated std. dev. $\sigma_c$ and true absolute error $|e_c|$, area under the curve (AUC), and computation time $t$ in seconds of CRM (ours), Log-Odds, Hilbert mapping and GPOM under different sensor model noise \revf{standard} deviations (right), where $s$ is the voxel size (\SI{0.05}{\meter}).}}%
      \label{tab:noise-results}
}
\end{floatrow}
\end{figure*}

The proposed method not only provides a more accurate map estimate but, more importantly, the uncertainty associated with the returned value. 
In doing so, it incorporates the full forward sensor model into the mapping process. 
As an example, it can distinguish between two voxels when they are both reported as almost free (\eg $ \hat{m}^{1}=\hat{m}^{2}=0.1 $) but have different confidence levels (\eg $ \sigma^{m^{1}}=0.01 $ and $ \sigma^{m^{2}}=0.2 $). 
This confidence level is a crucial piece of information for a motion planner. 
Obviously, the planner should either try to avoid $ m^{2} $ since the robot is not sure if $ m^{2} $ is actually risk-free (due to high variance) or it needs to take perceptual actions to obtain more measurements from $ m^{2} $ before passing through that part of the map.

The Log-Odds-based method stores only a single number per voxel in the map, namely the parameter of the Bernoulli distribution. 
One might try to utilize the variance of the Bernoulli distribution to infer the confidence in an Log-Odds-based map, but due to the incorrect assumptions made in the mapping process and since the Bernoulli distribution is a single-parameter distribution (mean and variance are dependent), the computed variance is not a reliable confidence source.	

It is important to note that a planner is generally able to cope with large errors \emph{if} there is a high variance associated with them. 
However, if the error is high but the mapping method is confident about its wrong estimate, motion planning becomes very risky and prone to failures. 

Consider a simple planning scenario in an unknown environment, where a quadrotor has to traverse the environment as fast as possible while ensuring a reasonable level of safety, as shown in Fig.~\ref{fig:planning}.
Using our CRM, one can predict the future variance of the occupancy of the path. 
Since our method yields consistent variance estimates, it enables uncertainty-aware planners to reliably reason about the information gain of future perceptual measurements.
Having this information opens new possibilities for uncertainty-aware planners such as incorporating safety-critical exploratory behavior into the fast navigation task.
We present a motion planning algorithm in~\cite{heiden2017planning} that leverages the CRM voxel representation. Equipped with a conservative optimization criterion that incorporates the estimated mean and variance of occupancy, the planner generates safe trajectories that are traversable at high velocities (cf. Fig.~\ref{fig:planning}).

\section{Results} \label{sec:results}
In this section, we demonstrate the performance of the proposed method and compare it with commonly used mapping methods. We start by studying the mapping error and consistency of the estimation process in a simulated scenarios \revf{and compare the results of CRM, Log-Odds, Hilbert and GP-based mapping approaches}. Subsequently, we analyze and compare the performance of Log-Odds mapping and CRM on real-world datasets. \rev{While the mapping results presented in this work are computed offline on a desktop computer, in~\cite{heiden2018fusion} we show results of CRM running on embedded systems fusing measurements from multiple depth sensors in real time.}

\subsection{Experimental Setup}
\label{sec:corridor}
Fig.~\ref{fig:ground-truth_map} shows a 2D ground-truth map which serves as a simulation environment for the following scenario. 
\rev{Each voxel is assumed to be a square with $s=\SI{0.05}{\meter}$ side length. 
The environment size is $2\times\SI{2}{\meter}$, consisting of $1,600$ voxels.} 
Each voxel is either fully occupied (shown in gray) or empty (white). 
The robot follows a trajectory \revf{of 23 waypoints} (red dots), as shown in Fig.~\ref{fig:ground-truth_map}, and takes measurements at these positions.

For the sensing system, we simulate a ranging sensor with $60$ omnidirectional depth sensors spanning a field of view of 360$^\circ$ and reaching up to a range of \SI{1}{\meter}.
Measurements are perturbed by zero-mean Gaussian noise with \SI{0.05}{\meter} \revf{standard} deviation, \ie one voxel length, unless denoted otherwise. 

\newcommand{\ScenarioFigSize}{1.09in}
\begin{figure*}[ht!]
	\centering
	\includegraphics[width=.75\textwidth]{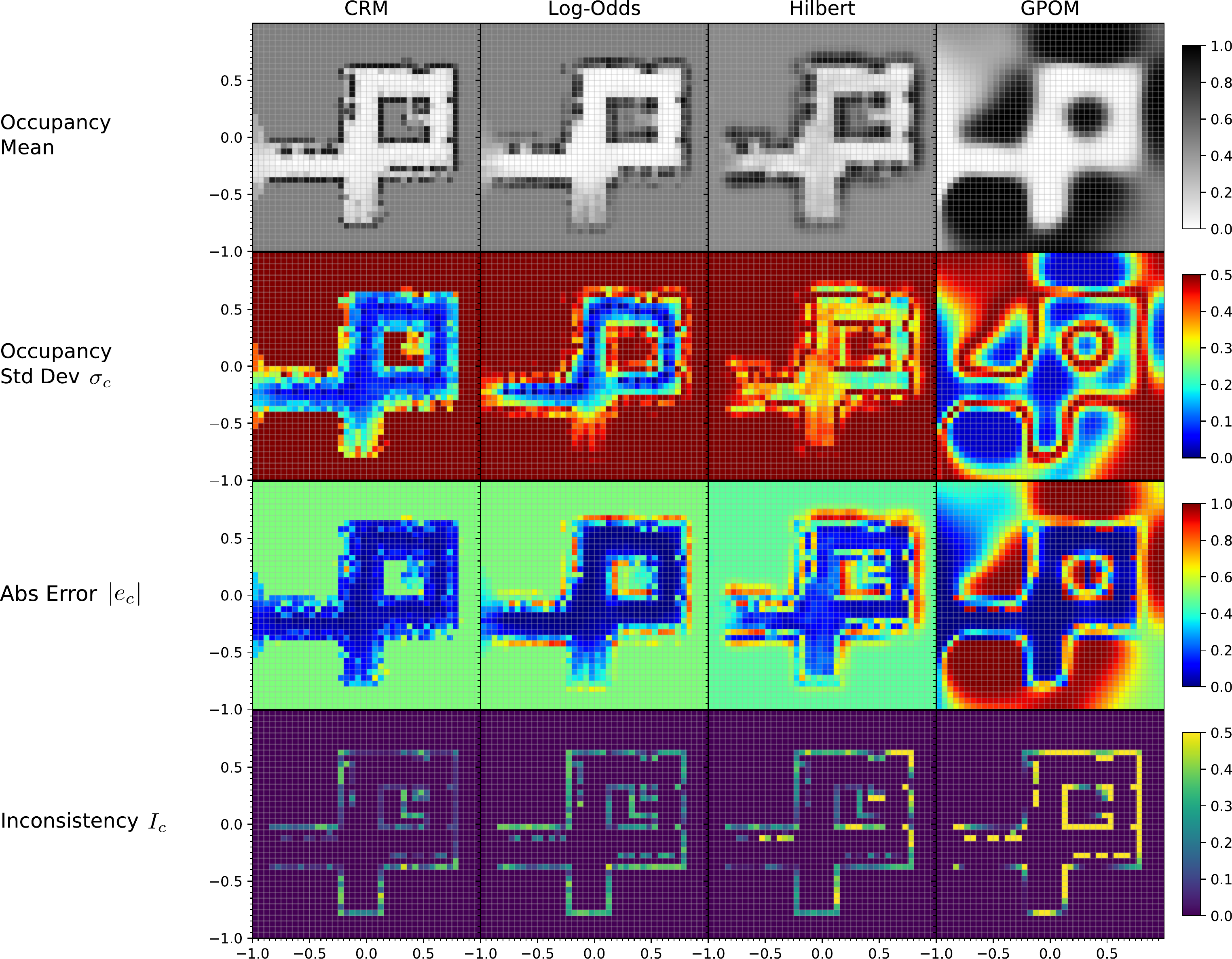}
	\caption{\rev{Mapping results for the simulation experiment from Sec.~\ref{sec:corridor} under $0.25s$ sensor noise \revf{standard} deviation using CRM, Log-Odds, Hilbert and GPOM (columns from left to right). First row: estimated mean occupancies. Second row: estimated \revf{standard} deviation $\sigma_c$. Third row: absolute error $|e_c|$. Fourth row: voxel-wise inconsistency $I_c$ with $\gamma = \nicefrac{1}{2}$ (cf. Eq.~\eqref{eq:inconsistencyMeasure}), \ie $I_c=\max(0, |e_c| - \nicefrac{1}{2} \sigma_c)$.
	Spatial units are in meters.}}
	\label{fig:example-map}
\end{figure*}

For Log-Odds mapping, we use an inverse sensor model (Fig.~\ref{fig:invSensorModel}) with parameters $ r_{ramp} = 0.1 $, $ r_{top}=0.1 $, $ q_{l}=0.45 $ and $ q_{h}=0.55 $ \rev{which yielded the lowest mean absolute error in our parameter grid search (cf. Fig.~\ref{fig:ismRuns})}.

\rev{Besides Log-Odds mapping as a grid-based approach, we compare our method against two other algorithms, namely Hilbert maps and Gaussian Process occupancy mapping (GPOM), that learn a map given samples of free and occupied areas in the environment. As shown in Fig.~\ref{fig:ground-truth_map}, we sample free space (blue) along the measurement rays by sampling from a uniform distribution of positions along the measurement ray, starting from the robot's position up until the measurement range has been reached or the map boundaries have been approached (whichever comes first). For measurements that reach an obstacle, we add the measurement's end point to the training dataset of occupied areas. Overall, the training dataset consists of $19,472$ samples \revf{in total}, where $1,302$ samples correspond to occupied areas and $18,170$ samples are labelled as free space.

For Hilbert Maps~\citep{ramos2016hilbert}, we use the Nystr\"om feature with RBF kernel parameter $\gamma=200$ and $2000$ components, \ie the subset of training samples used to construct the feature map, which empirically yielded the lowest mean absolute error. Since Hilbert maps perform logistic regression to predict the occupancy of the map, we treat the output as a Bernoulli random variable $p$ from which we compute the \revf{standard} deviation $\sqrt{p(1-p)}$, as in the case of Log-Odds mapping.

\revf{Based on experimental observations of the accuracy and speed trade-off of different variations of GP-based methods, we adopt Variational Sparse Dynamic Gaussian Process occupancy maps (VSDGPOM)~\citep{senanayake2017learning} as a representative GP-based method. Specifically,} VSDGPOM belongs to the class of dynamic GPOM which are able to continuously update the map as new sensor data becomes available and can therefore operate in dynamic environments. In contrast to traditional approaches, VSDGPOM uses techniques from variational inference to select inducing points in the data that accelerates the learning process in comparison to other GPOM approaches that use the entire dataset in the training process. As in~\citet{senanayake2017learning}, we process the samples from the training dataset at every measurement step using the Density-Based Spatial Clustering of Applications with Noise (DBSCAN) algorithm that automatically selects the clusters whose centroids serve as inducing points. We scale up the sampling location coordinates by $100$ during training and querying the GP as the generated maps appeared more distinct than maps obtained from samples at their original locations. For visualization and evaluation purposes, we project the output samples down to their original dimensions.}

\rev{The occupancy maps resulting from CRM, Log-Odds, Hilbert mapping and GPOM are shown in the first row of Fig.~\ref{fig:example-map}.}

\subsection{Simulation Experiments}
\label{sec:simulation}
\rev{In this experiment, we study the sensitivity of our method to different sensor noise intensities by varying the \revf{standard} deviation of the Gaussian noise over the values $\{0.25s, 0.5s, 1s, 2s, 3s\}$, where $s$ is the voxel size. 

We first focus on mapping accuracy and report the mean absolute error (MAE) for different sensor noise \revf{standard} deviations in the top rows of Table~\ref{tab:noise-results}.
For the same noise intensity, in most cases, CRM shows a smaller error compared to the mapping results from the other methods. Considering mapping algorithms as binary classifiers that predict whether a given location in space is free or occupied, we can leverage the area under the curve (AUC) of the receiver operating characteristic (ROC) as a performance measure. As shown in Table~\ref{tab:noise-results}, under modest sensor noise conditions, CRM consistently scores close to $1$, which represents a perfect classifier.}
\begin{figure}[ht!]
	\centering
	\subfloat{
	\includegraphics[width=\columnwidth,clip]{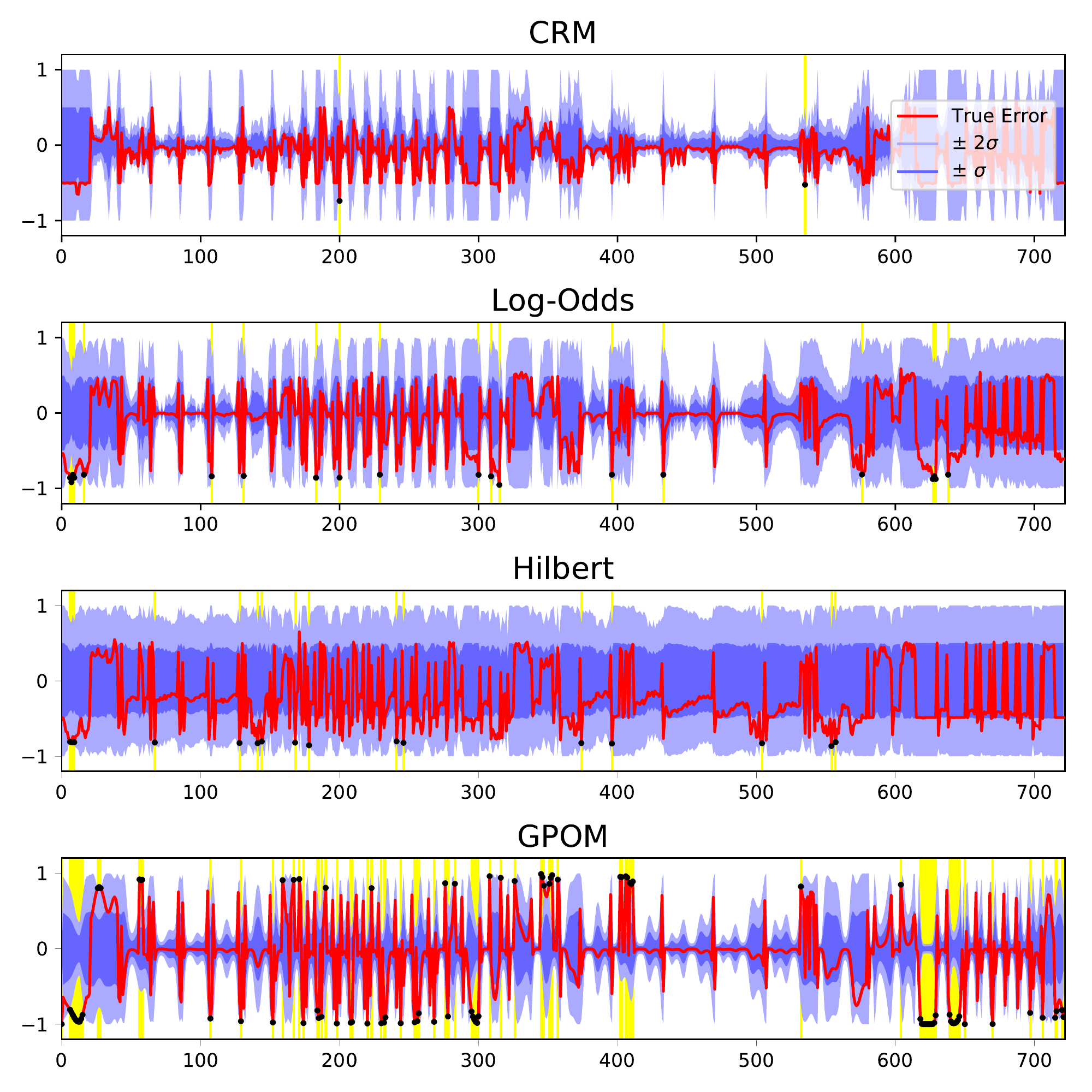}\vspace{-16pt}}
    \caption{\rev{Mapping error (red) and algorithmically computed \revf{standard} deviation (blue shades for $ 2\sigma $ (light) and $ \sigma $ (dark) confidence bounds) of all updated voxels in the simulation experiment from Sec.~\ref{sec:corridor}. Errors outside the $2\sigma$-interval are highlighted by black dots and yellow rectangles in the background. \revf{The abscissa corresponds to the voxel id.}}}
    \label{fig:bad_error_consistency}
\end{figure}

\begin{figure*}[ht!]
	\centering
	\subfloat{\includegraphics[height=3.5cm,trim=.6cm .6cm .6cm .6cm]{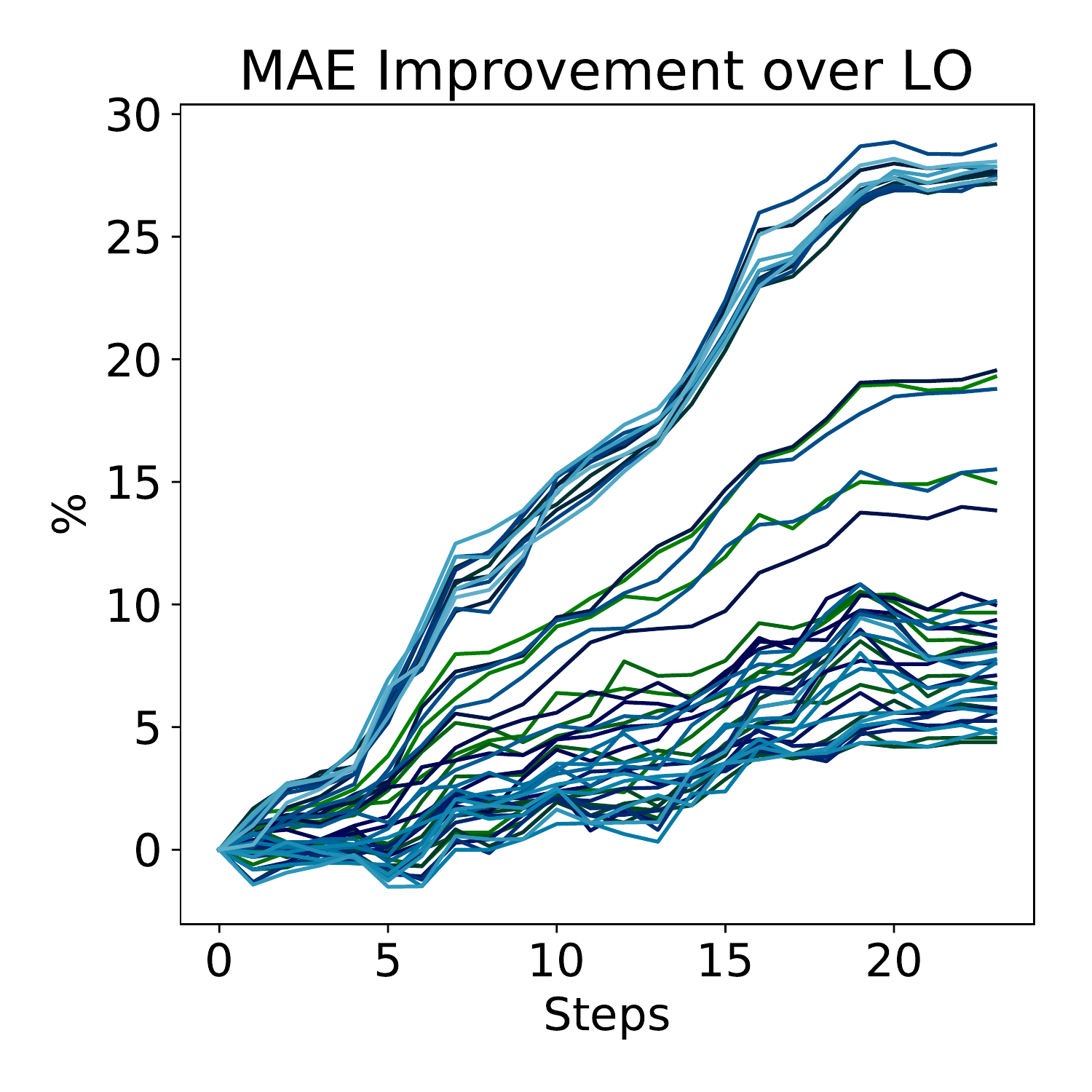}
	\label{fig:maeLogOddsImprovement}}
	\hspace{10pt}
	\subfloat{\includegraphics[height=3cm,trim=0cm .3cm 0cm 0cm,clip]{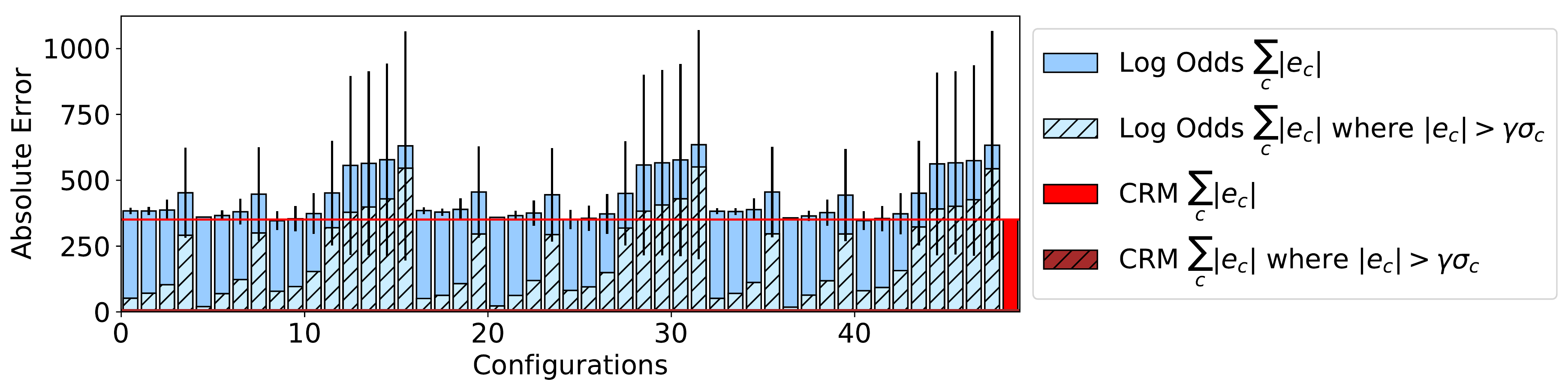}
	\label{fig:ismRuns}}
	\caption{\rev{Performance evaluation of CRM and Log-Odds with $48$ different ISM configurations for the experiment from \revf{Sec.~\ref{sec:simulation}}. \emph{Left:} Percentage improvement in mean absolute error (MAE) of CRM over Log Odds with different ISM \revf{configurations}. \emph{Right:} Comparison of the cumulative mapping error between Log-Odds mapping with different ISM \revf{configurations}, and CRM. The whiskers depict the inconsistency $I_c$ (Eq.~\ref{eq:inconsistencyMeasure}), the striped areas represent the portion of errors $|e_c|$ that are inconsistent with the reported \revf{standard} deviation (greater than $ \gamma\sigma_c $ for $ \gamma = 1.25 $).
	The proposed method results in a seven times more consistent mapping with one third of the critical errors due to overconfidence, compared to the best Log-Odds configuration. \revf{The red horizontal line shows the absolute error of CRM and is drawn to simplify the visual comparison between ISM configurations and CRM. Note that the striped red part and its associated whisker are too small, hence less visible.}}}
	\label{fig:barChart}
\end{figure*}

\begin{figure}[ht!]
	\centering
	\subfloat{\includegraphics[width=.49\columnwidth,trim=.6cm .6cm .6cm .6cm]{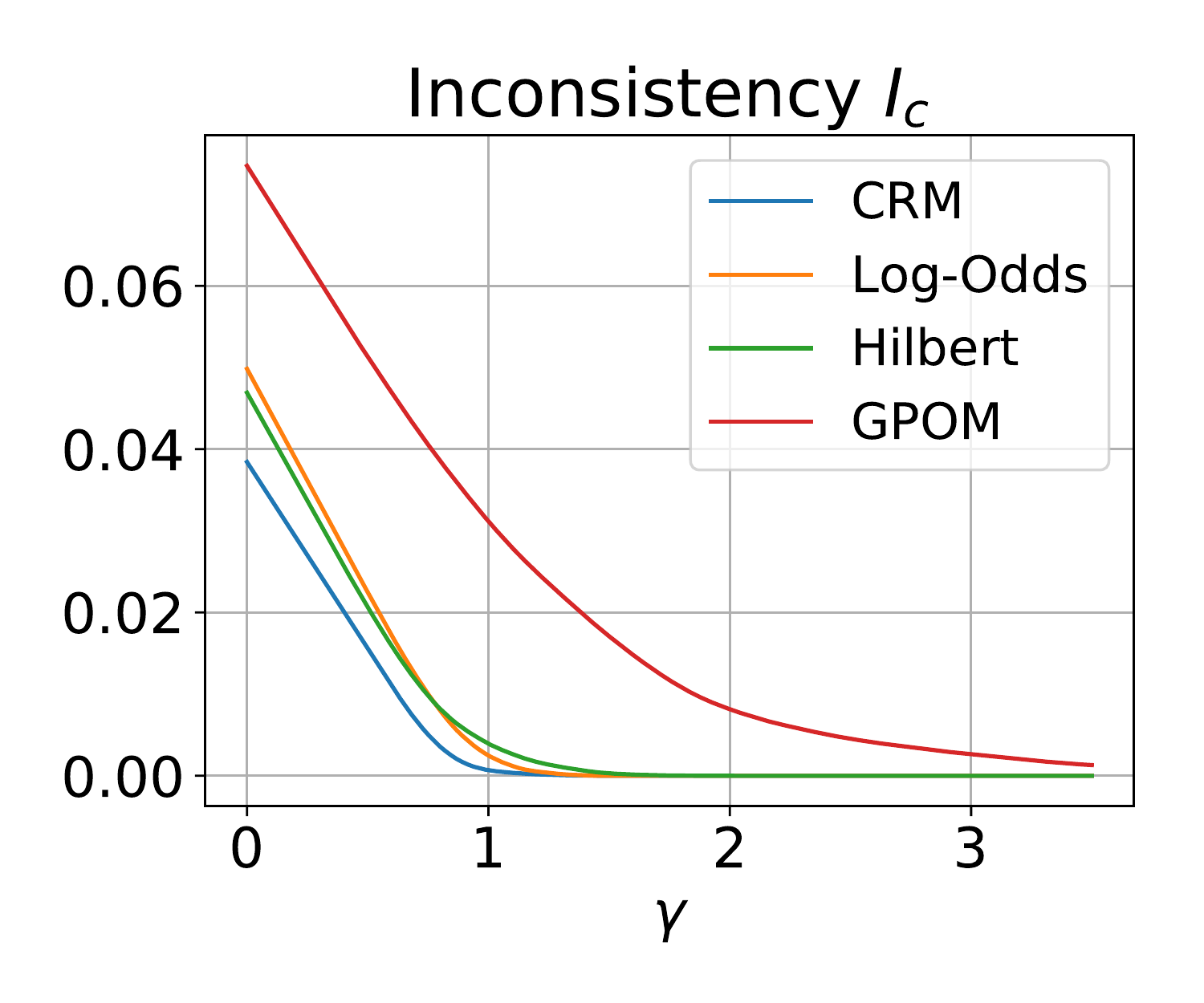}} \hfill
	\subfloat{\includegraphics[width=.49\columnwidth,trim=.6cm .6cm .6cm .6cm]{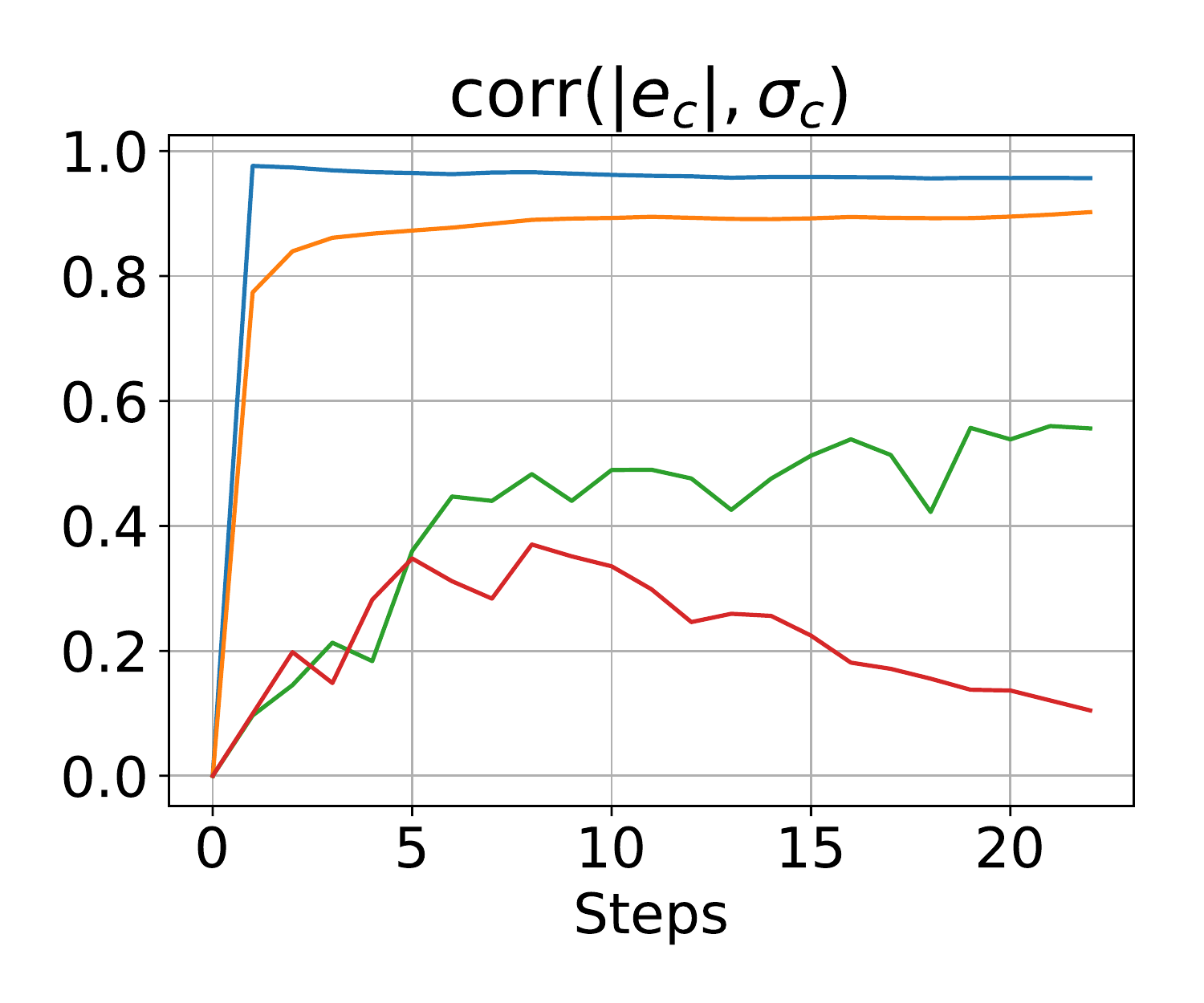}} \\
	\subfloat{\includegraphics[width=.49\columnwidth,trim=.6cm .6cm .6cm .6cm]{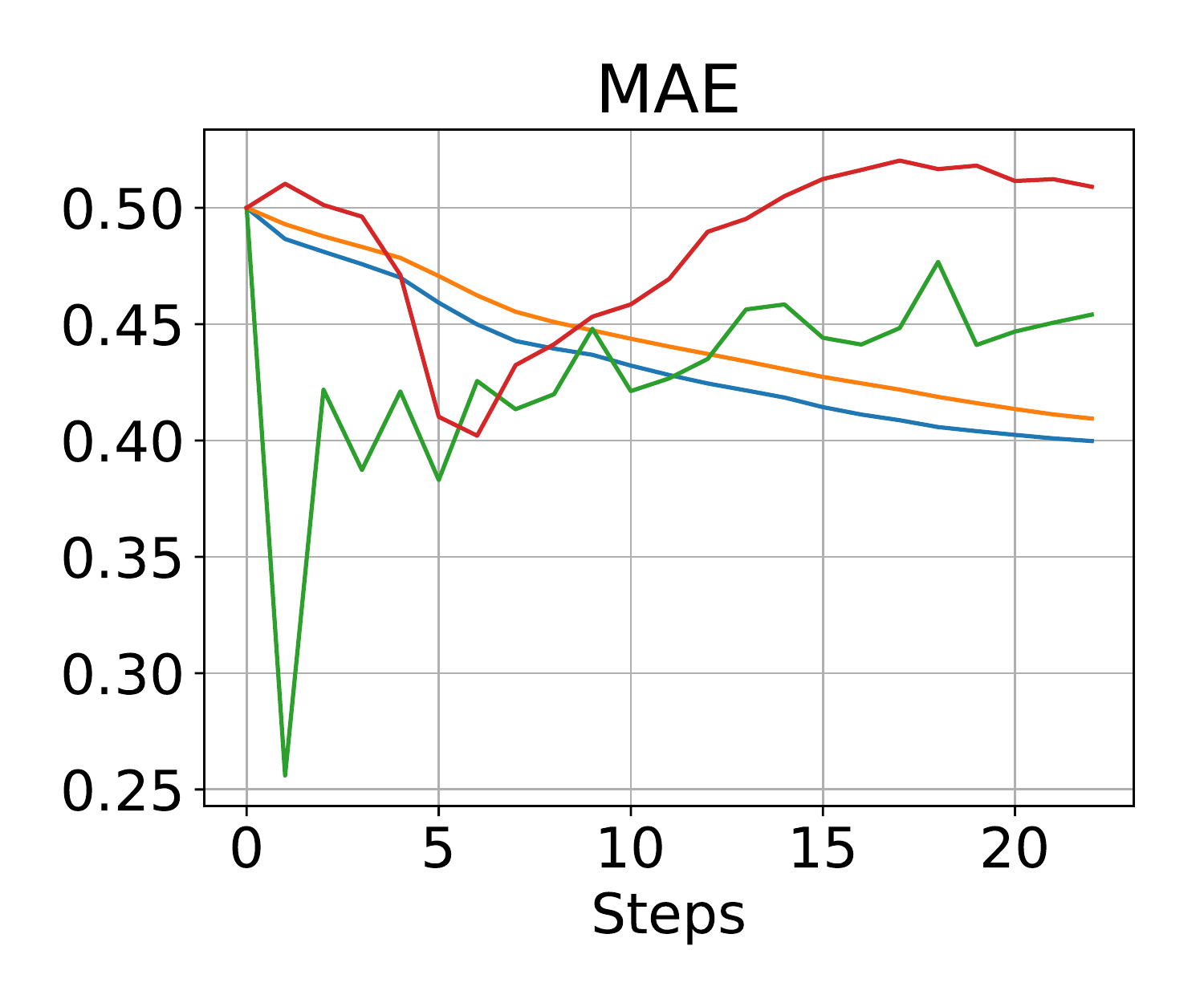}} \hfill
	\subfloat{\includegraphics[width=.49\columnwidth,trim=.6cm .6cm .6cm .6cm]{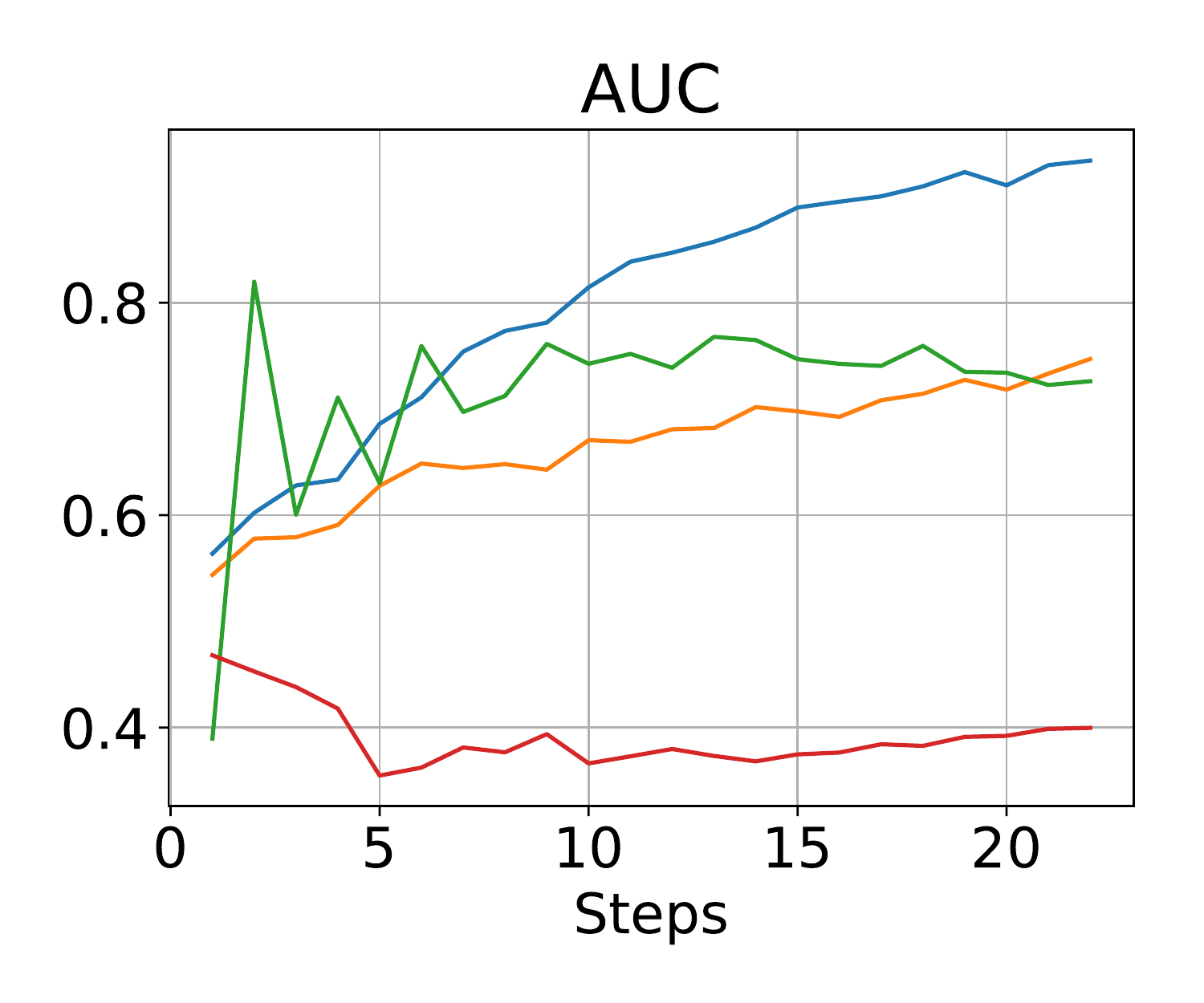}}
	\caption{\rev{Mapping statistics for the simulated experiment from Sec.~\ref{sec:corridor}: evaluation of inconsistency $I_c$ (Eq.~\ref{eq:inconsistencyMeasure}) over threshold values $\gamma$ (top left),  Pearson correlation coefficient between the reported \revf{standard} deviation and the absolute error (top right), mean absolute error (bottom left), and area under the curve (bottom right) over all voxels at a given map update step.}}
	\label{fig:inconsistency}
\end{figure}

\rev{
We visualize the occupancy mean, estimated \revf{standard} deviation, absolute error and inconsistency in Fig.~\ref{fig:example-map} with $0.25s$ sensor noise \revf{standard} deviation. In the visual comparison, CRM maps the obstacles most clearly while Log-Odds and Hilbert show slightly less distinct obstacle outlines. GPOM has a less distinct estimate of voxel occupancies, mapping spaces as free or occupied for which no measurement samples have been acquired.

In the third row of Fig.~\ref{fig:example-map}, we visualize the absolute error, ranging from $0$ (dark blue) to $1$ (dark red). CRM achieves the overall lowest error, while Hilbert and Log-Odds both exhibit higher deviations, in particular at obstacle locations. The less distinct occupancy estimates from GPOM affect the accuracy, leading to areas in the map with high errors (note that the maximum possible error is $1$).
}

As discussed in Sec.~\ref{sec:planning}, reducing the map error is only an ancillary benefit of the proposed method. 
The main objective of CRM is to provide a consistent confidence measure which is particularly important for planning purposes. 
A planner might be able to handle large errors as long as the mapping method indicates that the estimates are unreliable (\eg via their variance). 
However, if the filter returns a wrong estimate with low variance, \ie it is overconfident, then the planner is prone to yield unsafe results. 

To quantify the \emph{inconsistency} between the error and reported variances, we use the following measure:
\begin{align} %
I_c = \sum_{c}\max(0, |e_{c}|-\gamma\sigma_{c}),%
\label{eq:inconsistencyMeasure} %
\end{align} %
\noindent \revf{where} $ e_c $ and $ \sigma_{c} $ denote the estimation error (\ie the difference between the true and estimated voxel occupancy) and the estimated \revf{standard} deviation of voxel $ c $, respectively. $ \gamma $ decides what level of error is acceptable.
\rev{The ramp function \mbox{$ \max(0,x) $} ensures that only inconsistent voxels (with respect to $ \gamma\sigma_c $) contribute to the summation and therefore penalizes overconfident estimates.}

\rev{As can be seen in the bottom row of Fig.~\ref{fig:example-map}, where voxels of high inconsistency appear brighter, the GPOM-based approach tends to be overly confident in false estimates, in contrast to the other methods. Areas in the map that have not been observed are assigned high confidence values, rendering the overall certainty estimation of GPOM unreliable. Log-Odds tends to be less consistent at small obstacles which can represent dangerous scenarios for a confidence-aware motion planner. Although Hilbert predicts overall high \revf{standard} deviation at most voxels, the inconsistency is significant at many narrow passages that typically pose the highest risk to autonomous robot operations. CRM achieves the most consistent result overall throughout all noise intensities, as reported in Table~\ref{tab:noise-results}.}

\rev{While CRM exhibits the lowest inconsistency according to the $I_c$ measure for $\gamma=\nicefrac{1}{2}$, we investigate the response over a range of threshold values (setting $\gamma$ to values from $0$ to $3$) in Fig.~\ref{fig:inconsistency} (top left) for a sensor noise \revf{standard} deviation of $1s$. The $I_c$ measure only captures the error that lies outside the $\gamma\sigma_c$ interval and therefore rewards mapping algorithms that underestimate the confidence, \ie where $\sigma_c$ is very high. As shown in Fig.~\ref{fig:example-map} (second row, third column), Hilbert, for example, produces high uncertainty estimates throughout the map, even in areas where many training samples exist. Therefore, we turn to another metric that measures how predictive the estimate \revf{standard} deviation is for the true absolute error. The Pearson correlation coefficient (PCC) expresses the linear correlation between estimates as a value ranging between $-1$ and $+1$, where $1$ indicates perfect correlation, $0$ no correlation, and $-1$ represents total negative correlation. As shown in Table~\ref{tab:noise-results}, CRM outperforms all other mapping approaches significantly throughout all sensor noise intensities, yielding near-optimal linear correlation between the predicted uncertainty and the true absolute error.}

\rev{The higher consistency in the CRM estimate is further validated when we visualize the true error and the $\sigma_c$ and $2\sigma_c$ confidence bounds for every voxel that has been affected by the depth measurements (cf. Fig.~\ref{fig:ground-truth_map}) in Fig.~\ref{fig:bad_error_consistency}.}
As shown in the top row, the $ \sigma $-bound from CRM grows and shrinks in a consistent manner with the error, and behaves as a reliable confidence interval. 
\rev{In contrast, Log-Odds (second row), Hilbert (third row) and GPOM \revf{(fourth row)} exhibit increasingly more erroneous occupancy estimates \textit{at which the variance is very low}, which pose a significant challenge to a confidence-aware motion planner.}

\rev{Besides the accuracy and consistency of the resulting map, the computation time plays a deciding factor when deploying the mapping algorithms on resource-constrained robotic systems. In Table~\ref{tab:noise-results}, we report the time $t$ (in seconds) it takes to generate each of the maps. CRM and Log-Odds operate at similar speeds, with Log-Odds outpacing CRM by \SI{3}{\ms}. More significant differences appear in comparison to Hilbert and GP mapping whose computation times are three and up to five orders of magnitudes slower, respectively.

\revf{It should be noted that our implementation of Log-Odds is in C++. The CRM results are based on non-optimized C++ code. Hilbert maps and GPOM are implemented in Python interfacing with optimized C/C++ backends. Specifically, in the case of Hilbert mapping, we use operations from the scikit-learn and NumPy libraries that internally execute highly optimized, native code written in C/C++. In the case of GPOM, our implementation is based on TensorFlow which similarly is a Python front-end to a highly optimized C++ backend that uses NVIDIA CUDA to execute the operations on the GPU. Given these points, we believe the overall trend of the results is not biased by implementation details.}

For the evaluation, we execute the implementations on an Intel Core i7 8700 CPU (\SI{3.7}{\giga\hertz}), while for GPOM, we run a GPU-based implementation on an NVIDIA GeForce GTX 1080 Ti GPU. Besides training Hilbert and GPOM, it takes significant computational effort to query the maps~\citep{OCallaghan2012-IJRR,ramos2016hilbert}, which we did not take into consideration in our timing comparison. Given the computational limitations, for further experiments with more measurements and experiments where we generate 3d maps, we continue the comparison with Log-Odds only.}

Since the inverse sensor model (ISM) is typically hand-engineered for a given sensor and a given environment, we sweep over a set of ISM parameters to compare the performance with our proposed method. Following the generic form of the ISM in Fig.~\ref{fig:invSensorModel}, \rev{we create 48 models by \textit{(i)} setting the $ q_{h}-0.5=0.5-q_{l} $ to $\{0.05,0.2,0.4\}$, \textit{(ii)} setting $ r_{ramp} $ to $\{0.03, 0.05, 0.1, 0.3\}$ and \textit{(iii)} setting $r_{top}$ to $\{0.03, 0.05, 0.1, 0.3\}$.}
The results reported in Fig.~\ref{fig:barChart} show that CRM consistently produces a smaller absolute error than Log-Odds over a variety of ISM \revf{configurations}, as well as a smaller portion of error that is inconsistent according to the $I_c$ measure (Eq.~\ref{eq:inconsistencyMeasure}).

\revf{Fig.~\ref{fig:inconsistency} summarizes the results from our simulation experiment described in Sec.~\ref{sec:simulation} where we evaluate CRM, Log-Odds, Hilbert mapping and GPOM on four different metrics. We visualize the inconsistency measure over different threshold values (top left). In the top right subplot, we compare the evolution of the average Pearson correlation coefficient of all voxels between the true absolute error $|e_c|$ and the estimated std deviation $\sigma_c$. Throughout the 23 map updates from the measurements taken at the locations shown in Fig.~\ref{fig:ground-truth_map} (red dots), CRM estimate demonstrates a high correlation (close to 1), closely followed by Log-Odds. Hilbert mapping reaches a correlation of $\sim$0.58 by the end of the run, while GPOM degrades in correlation midway through the experiment, which indicates that the predicted std deviation does not reflect the actual mapping error in this case. Such behavior can be observed in Fig.~\ref{fig:example-map} in the second and third row: while the std deviation of GPOM (last column) is close to zero (dark blue), the true absolute error reaches its maximum of 1 (dark red) in various parts of the map. The mean absolute error shown in Fig.~\ref{fig:inconsistency} (bottom left) steadily decreases for CRM and Log-Odds over the course of the execution, while the evolution of this metric fluctuates around 0.5 for Hilbert and GPOM. Comparing the mapping methods based on the area-under-the-curve (AUC) measure (bottom right) exhibits more significant differences between different mapping methods: In this example, CRM has a higher accuracy (close to 1) compared to the other methods. Log-Odds and Hilbert mapping approach a similar final AUC of approximately 0.75, and GPOM converges to 0.4.}

\subsection{Real Data Experiments}
\label{subsec:realdata}

\begin{table}[]
    \centering
    \begin{tabularx}{\columnwidth}{ X r r }
    \toprule
	Inconsistency $I_c$ & \multicolumn{1}{c}{Log-Odds} & \multicolumn{1}{c}{CRM} \\
    \midrule
	\tt albert-b-vision & 984.777 & \bf 716.211 \\
	\tt ICL-NUIM & 2,871.180 & \bf 1,307.793 \\
	\tt EuRoC MAV & 645.750 & \bf 434.598 \\
	\tt USC Intel Drone & 6,344.294 & \bf 1,196.418 \\
	\bottomrule
    \end{tabularx}
    \caption{Map inconsistencies $I_c$ (with $\gamma=2$ in Eq.~\ref{eq:inconsistencyMeasure}) of Log-Odds and CRM, evaluated on the real-world datasets.}
    \label{tab:real_inconsistent}
\end{table}

In addition to the simulation experiments, we evaluate the proposed method on commonly-used real-world datasets as well as our own dataset captured by a stereo camera on a physical robot. Table~\ref{tab:real_inconsistent} summarizes the most important result of this section over four different real-world datasets. In the following subsections, we will discuss these datasets and the evaluation results in further detail.

\subsubsection{Radish Dataset \texttt{albert-b-vision}}~\\

\newcommand{\ThreeFigSize}{2.08in}
\begin{figure*}[h!]
	\centering
	\subfloat{\includegraphics[width=\ThreeFigSize]{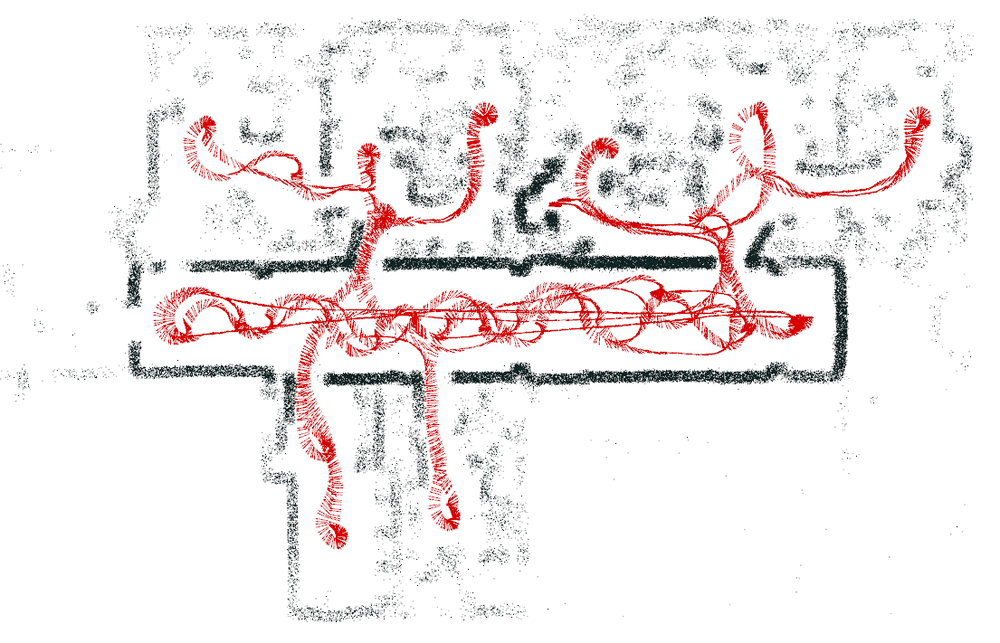}}
	\hspace{1pt}
	\subfloat{\includegraphics[width=\ThreeFigSize]{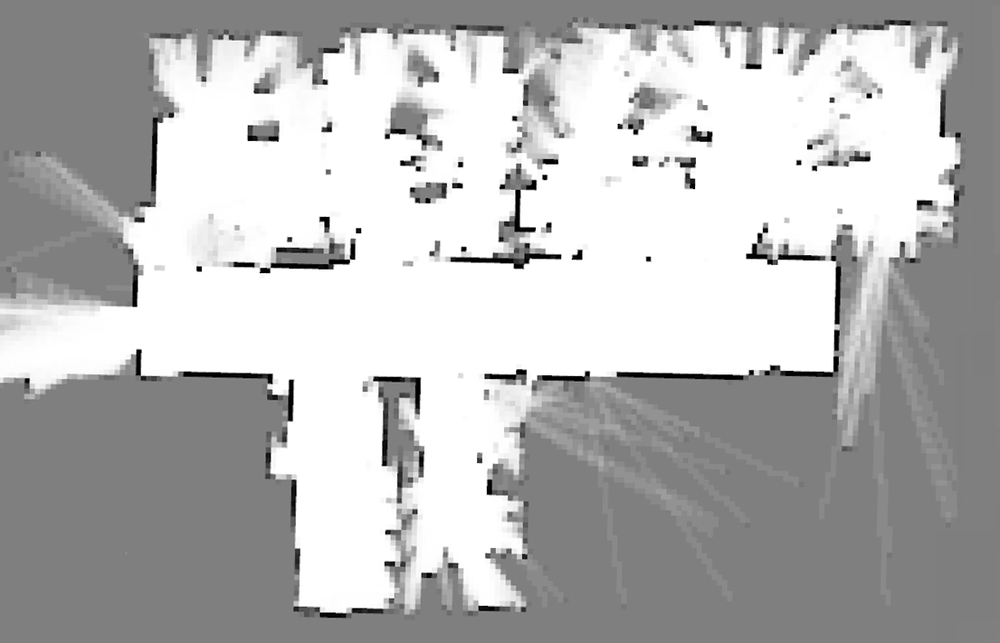}}
	\hspace{1pt}
	\subfloat{\includegraphics[width=\ThreeFigSize]{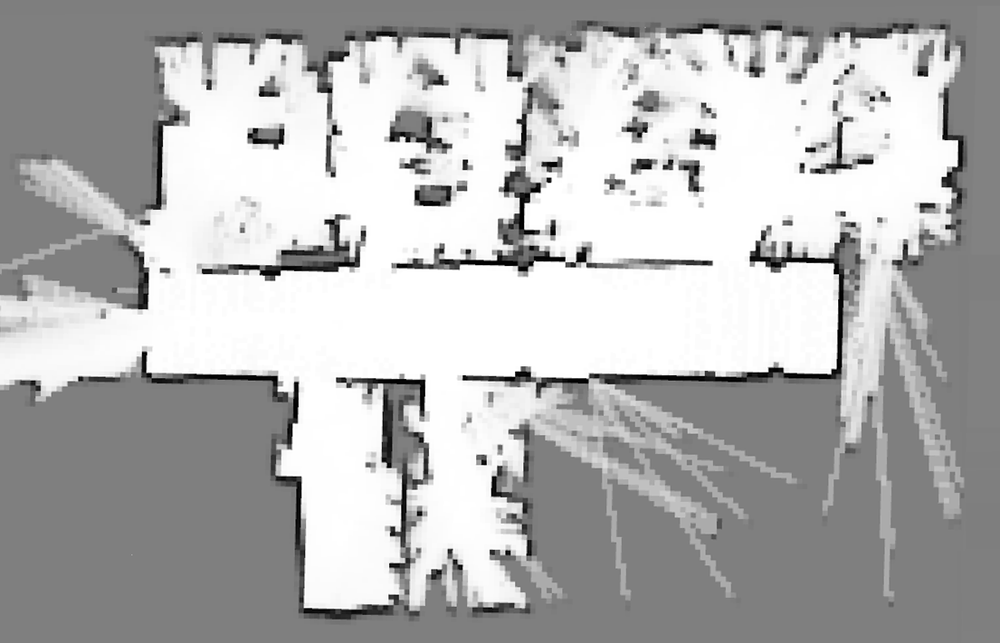}}
	\caption{\textit{Left:} end points of laser range measurements (black dots) and robot pose trajectory (red arrows) from the \texttt{albert-b-laser-vision} dataset. \textit{Center:} Log-Odds mapping results. \textit{Right:} mapping results from the proposed method. As it can be seen, CRM offers a more complete and more accurate map.}
	\label{fig:albert-carmen}
\end{figure*}

\begin{figure*}[ht!]
	\centering
	\subfloat{\includegraphics[height=3.5cm,trim=.6cm .6cm .6cm .6cm]{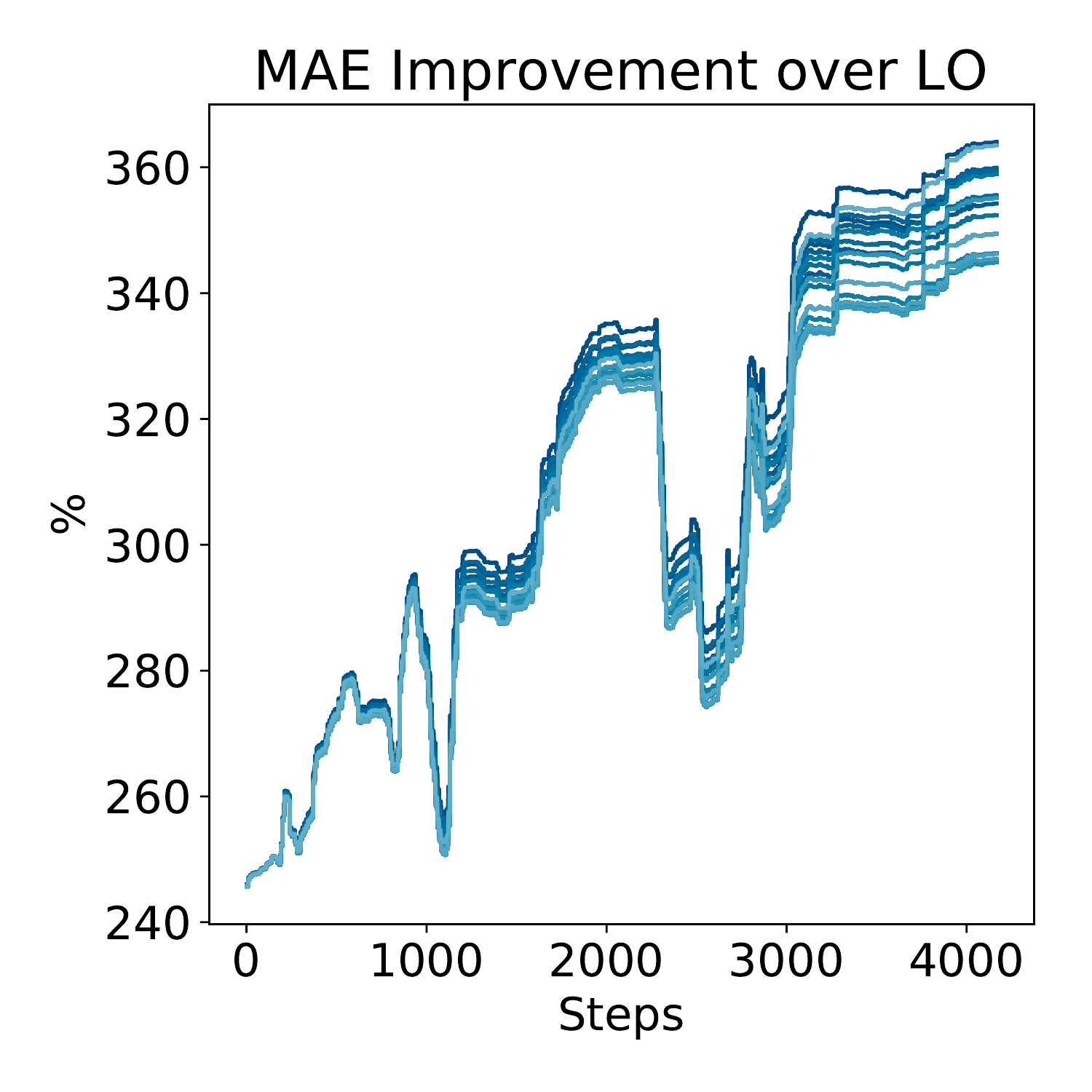}
	\label{fig:albert_maeLogOddsImprovement}}
	\hspace{10pt}
	\subfloat{\includegraphics[height=3cm,trim=0cm .3cm 0cm 0cm,clip]{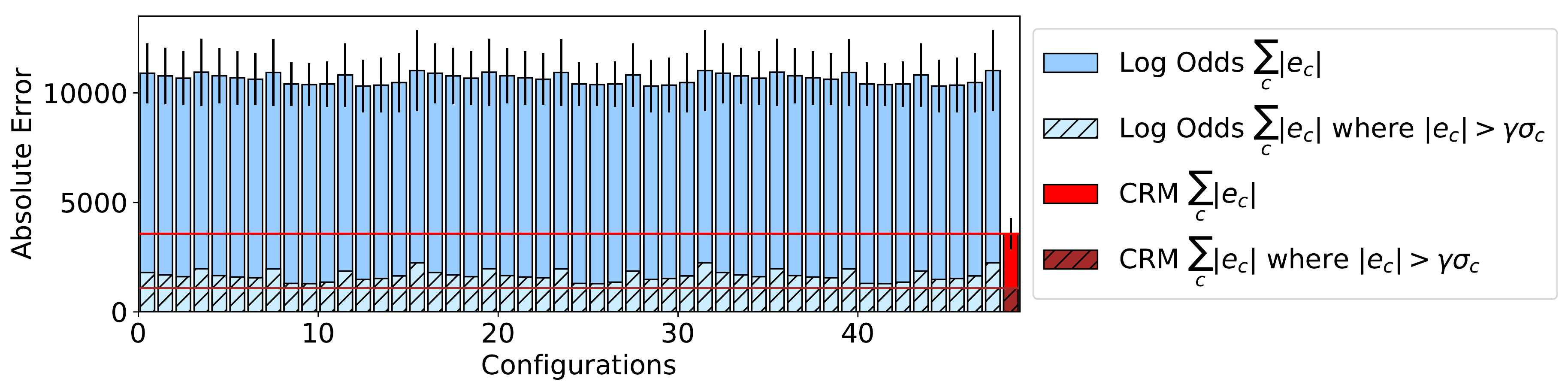}
	\label{fig:albert_ismEvaluation}}
	\caption{\textit{Left:} percentage improvement in mean absolute error (MAE) of CRM over Log Odds with 27 different ISM \revf{configurations} on the \texttt{albert-b-laser-vision} dataset. \textit{Right:} comparison of cumulative mapping error between Log-Odds with different ISM configurations and our method. The whiskers depict the inconsistency $I_c$ (Eq.~\ref{eq:inconsistencyMeasure}), the striped areas represent the portion of errors $|e_c|$ that are inconsistent with the \revf{estimated} std (greater than $ \gamma\sigma_c $ for $ \gamma = 1.25 $).}
	\label{fig:albert-comp}
\end{figure*}

In this subsection, we evaluate the proposed method on the dataset \texttt{albert-b-laser-vision} from the Robotics Data Set Repository (Radish)~\citep{Radish}.

The dataset contains more than $4,000$ sampling poses at which $180$ laser range measurements were recorded over a field of view of $180^\circ$ (cf. Fig.~\ref{fig:albert-carmen}). A ground-truth map over a $240\times240$ grid of voxels with \SI{0.125}{\meter} side length has been computed using the Log-Odds mapping method from all of the measurements. In the mapping evaluation, only every tenth measurement from each sampling (\ie 10\% of the data) has been given as input to Log-Odds and CRM.

Figure~\ref{fig:albert-comp} shows the comparison results between the proposed method and Log-Odds, given the ground-truth map provided by the dataset. For the Log-Odds method, we consider 48 different ISM model parameters, as described in Sec.~\ref{sec:simulation}. The proposed method yields an improvement in mapping accuracy, \ie lower mean absolute error (MAE), of more than three times compared to the best performing ISM model (Fig.~\ref{fig:albert_maeLogOddsImprovement}). The inconsistency (Eq.~\ref{eq:inconsistencyMeasure}) of the final Log-Odds mapping is $984.777$ compared to an $I_c$ of $716.211$ in the CRM, while the fraction of erroneous predictions with overconfident variance estimates is 20\% higher in the Log-Odds mapping compared to our method. Over 1,000 measurements, the average run-time of Log-Odds is \SI{1.362}{\second} compared to \SI{4.474}{\second} of CRM.

\subsubsection{ICL-NUIM RGB-D Benchmark}~\\

\begin{figure}
	\centering\vspace{-1.5em}
	\includegraphics[width=.75\columnwidth]{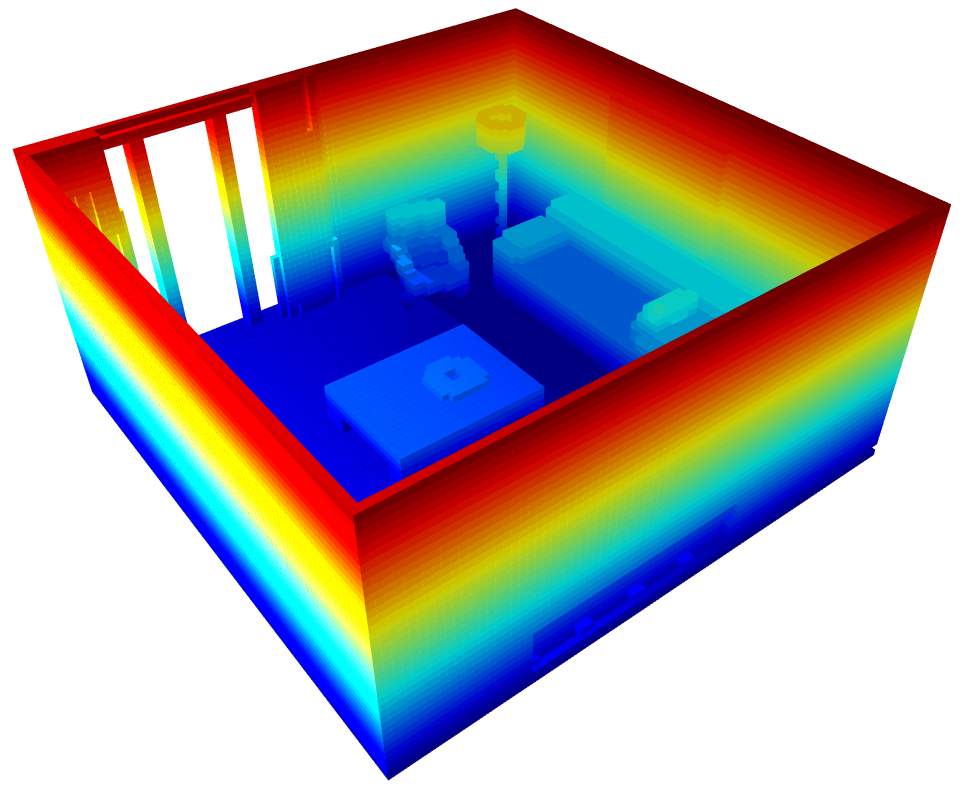}
	\caption{\rev{Ground-truth map for the \emph{ICL-NUIM living room} dataset, sliced along the $y$- and $z$-axes, and colored by voxel height for improved visibility.}}
	\label{fig:iclnuim_truth}
\end{figure}

\begin{figure}
	\centering
	\subfloat{\includegraphics[width=.95\textwidth]{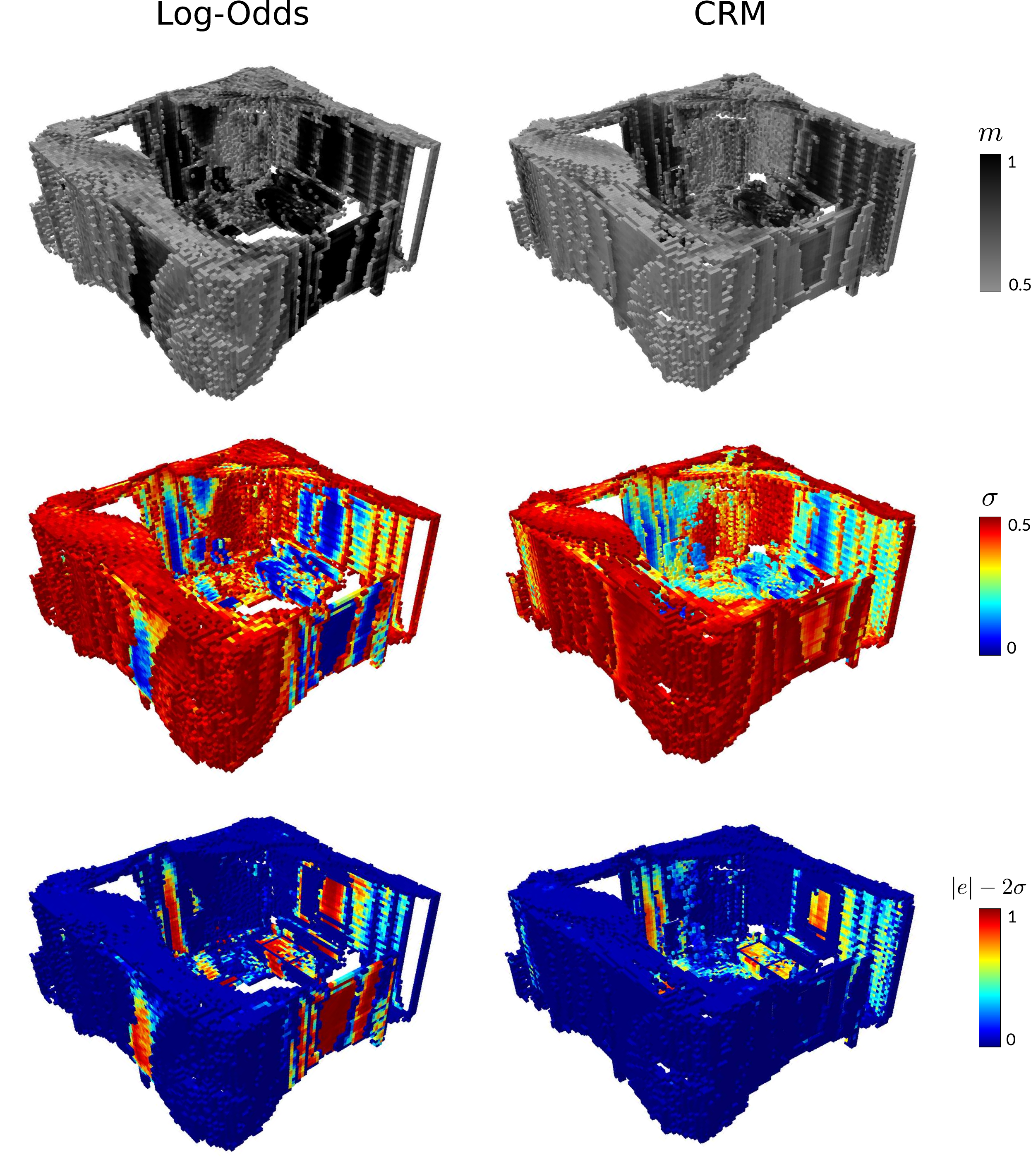}}
	\caption{\rev{Occupancy grids computed via Log-Odds (left column) and CRM (right column) resulting from the ICL-NUIM living room dataset (Trajectory-1): (top row) voxel mean occupancy, (middle row) estimated \revf{standard} deviation, and (bottom row) inconsistency $I_c$ with $\gamma=2$ (Eq.~\ref{eq:inconsistencyMeasure}). Only voxels with an estimated occupancy greater than 0.5 are shown. As can be seen at the bottom left, the Log-Odds estimates are considerably less consistent with many errors lying outside the $2\sigma$ confidence interval.}}
	\label{fig:iclnuim_maps}
\end{figure}

\begin{figure}
	\centering
	\includegraphics[width=\columnwidth]{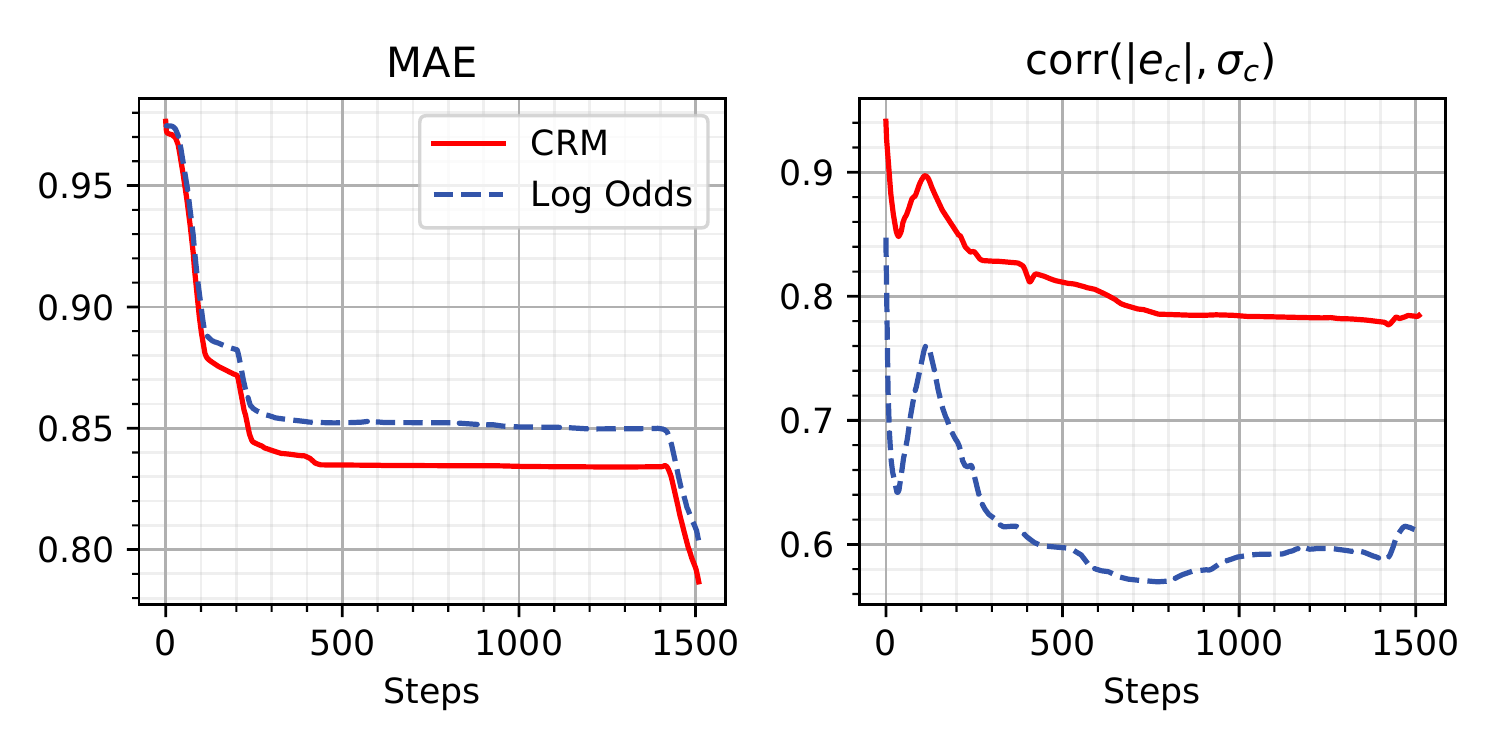}
	\caption{Mapping evaluation results of Log-Odds and CRM on the ICL-NUIM living room (Trajectory-1) dataset.
	\textit{Left:} evolution of the mean absolute error (MAE) between the estimated occupancy and ground-truth map.
	\textit{Right:} Pearson correlation coefficient between the true absolute error $|e_c|$ and the estimated voxel \revf{standard} deviation $\sigma_c$.}
	\label{fig:iclnuim_stats}
\end{figure}

\begin{figure}[ht!]
	\centering
	\subfloat{\includegraphics[width=\columnwidth]{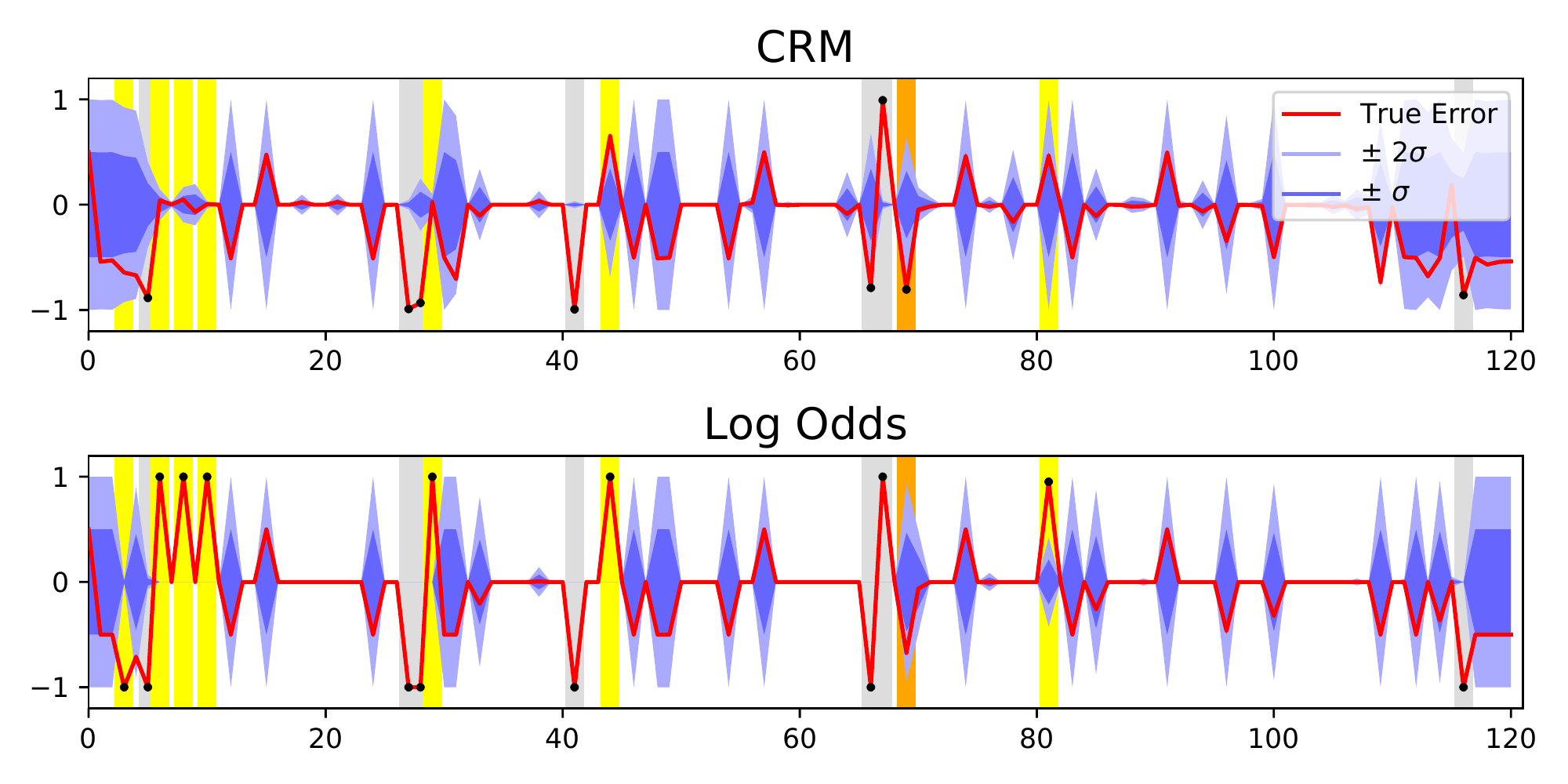}}
	\caption{True mapping error (red) and estimated \revf{standard} deviation (blue shades for $ 2\sigma $ (light) and $ \sigma $ (dark) confidence bounds) over a set of voxels for the ICL-NUIM dataset, with inconsistencies (black dots) where the error is outside the $2\sigma$ confidence bound. \revf{The abscissa corresponds to the voxel id.} Highlighted are instances where only Log-Odds (yellow), only CRM (orange) and both methods (light grey) are inconsistent. Many inconsistent occupancy estimates in the Log-Odds map are severe cases of overconfidence: some voxels with an estimated \revf{standard} deviation close to zero exhibit an error in estimated mean occupancy of almost one.}
	\label{fig:iclnuim_voxels}
\end{figure}

The ICL-NUIM dataset~\citep{handa2014iclnuim} contains synthetic depth measurements in a photo-realistically rendered indoor environment (see ground-truth map in Fig.~\ref{fig:iclnuim_truth}). The camera motion resembles that of a human operator since the trajectory has been motion-captured from a real human holding a camera. We use Trajectory-1 of the living room dataset with perfect depth information to compare CRM against Log-Odds-based mapping. Fig.~\ref{fig:iclnuim_maps} shows the result of this comparison for map error, \revf{standard} deviation, and estimation consistency. Fig.~\ref{fig:iclnuim_stats} captures the evolution of map error and correlation metric over time as new perception data arrive.

Given a voxel size of \SI{6.25}{\cm}, the grid consists of $608,256$ voxels in total. As can be seen on a subset of the updated voxels in Fig.~\ref{fig:iclnuim_voxels}, the estimated \revf{standard} deviation in CRM provides a more reliable confidence measure that better bounds the error compared to the Log-Odds map, resulting in a correlation between true absolute error and estimated \revf{standard} deviation of approximately $0.78$ and $0.61$ for CRM and Log-Odds, respectively. Table~\ref{tab:real_inconsistent} summarizes the overall map inconsistency $I_c$.

\subsubsection{EuRoC MAV Dataset}
\begin{figure}[h!]
	\centering
	\subfloat{\includegraphics[width=.75\columnwidth]{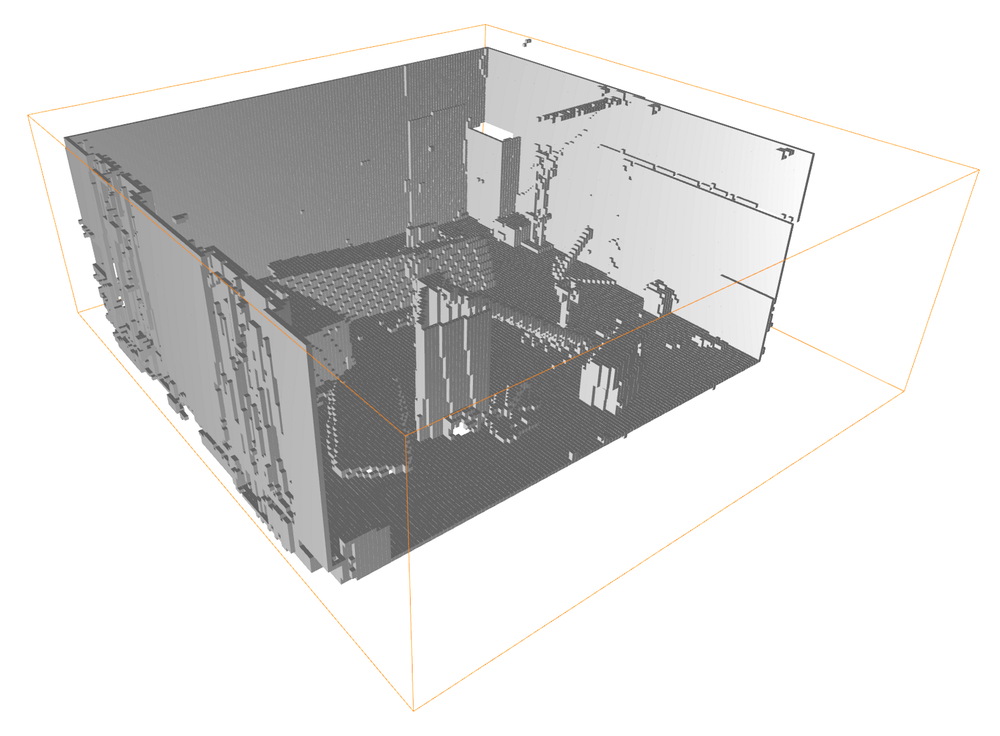}}
	\caption{Ground-truth map for the \emph{EuRoC MAV Vicon room~2} dataset, sliced along the $y$-axis for improved visibility.}
	\label{fig:euroc_truth}
\end{figure}

~\\
We evaluate CRM and Log-Odds mapping on the Vicon room dataset \texttt{V2\_03\_difficult} from the EuRoC MAV Dataset by~\citet{Burri25012016} (ground-truth map given in Fig.~\ref{fig:euroc_truth}) which presents a more challenging scenario for the perception system as the drone flies at higher speeds resulting in motion blur.

\begin{figure}
	\centering
	\subfloat{\includegraphics[width=.95\textwidth]{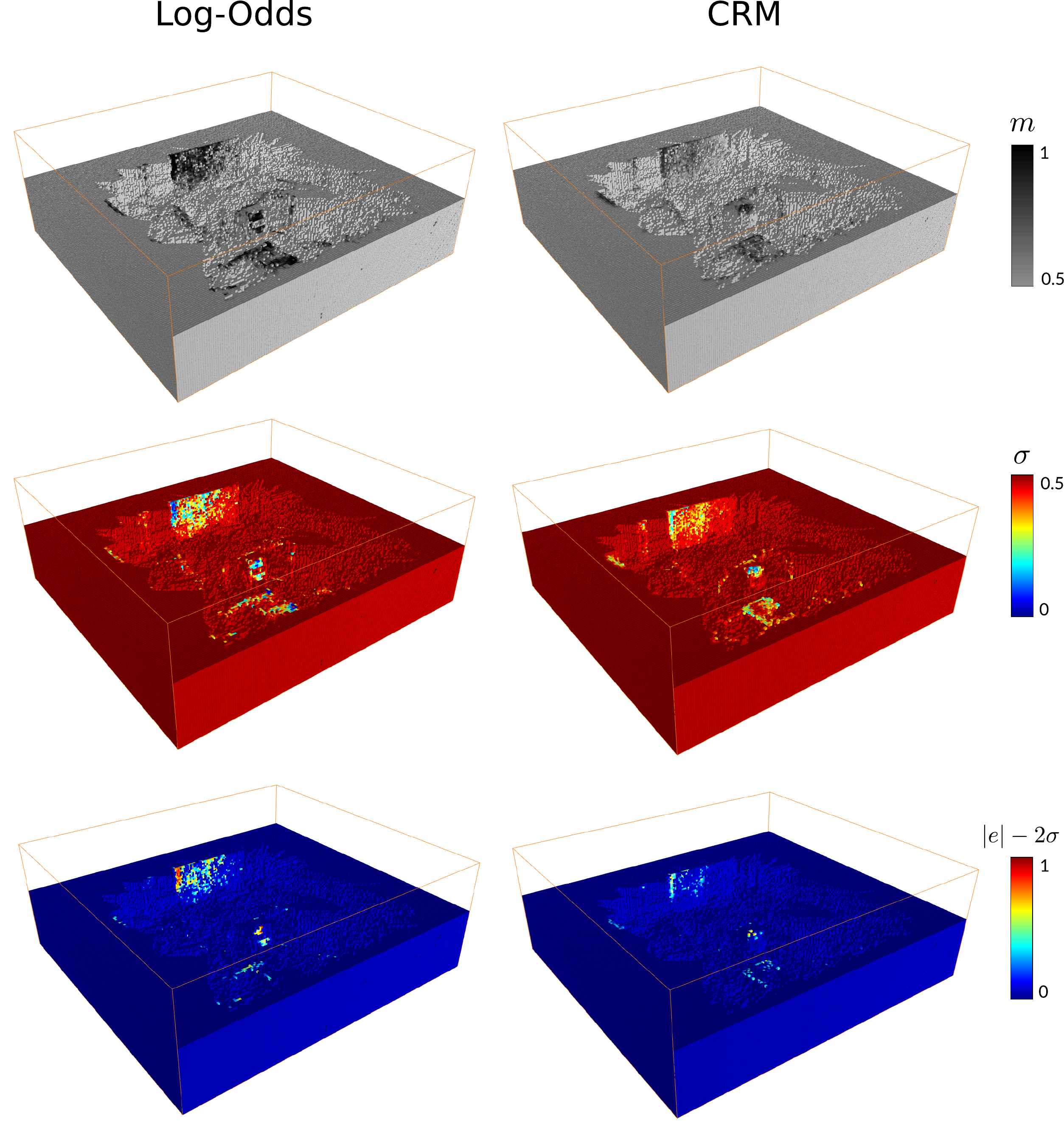}}
	\caption{\rev{Occupancy grids computed via Log-Odds (left column) and CRM (right column) resulting from the EuRoC MAV dataset (\texttt{V2\_03\_difficult}). We compare the results in terms of voxel mean occupancy (top row), estimated \revf{standard} deviation (middle row) and inconsistency $I_c$ (bottom row) with $\gamma=2$ (Eq.~\ref{eq:inconsistencyMeasure}).} The maps are sliced at approximately half the dimension along the $z$ axis for better visibility, the orange bounding boxes outline the original map dimensions. Only voxels with an estimated occupancy greater than \emph{or equal to} $0.5$ are shown.}
	\label{fig:euroc_maps}
\end{figure}

\begin{figure}
	\centering
	\includegraphics[width=\columnwidth]{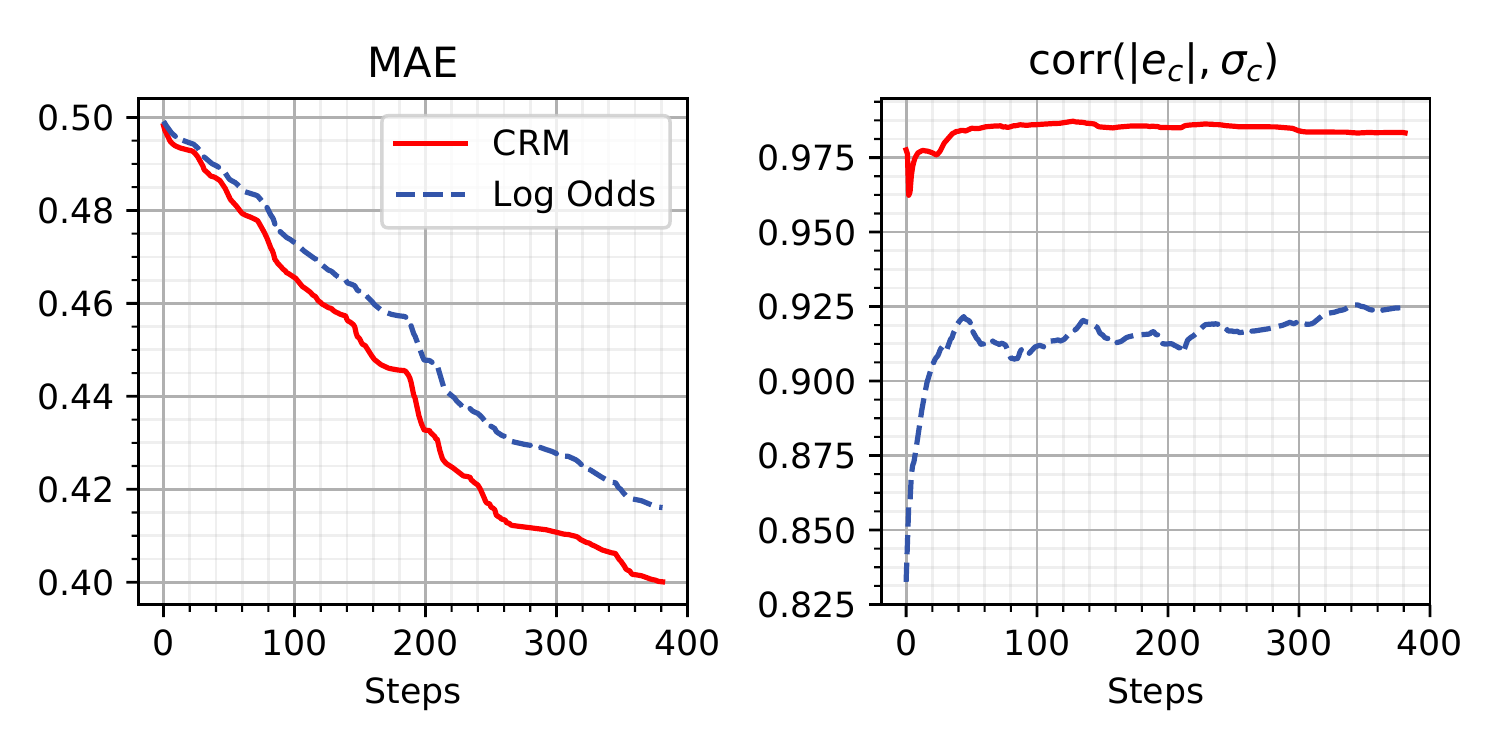}
	\caption{Mapping evaluation results of Log-Odds and CRM on the EuRoC MAV \texttt{V2\_03\_difficult} dataset.
	\textit{Left:} evolution of the mean absolute error (MAE) between the estimated occupancy and ground-truth map.
	\textit{Right:} Pearson correlation coefficient between the true absolute error and the estimated voxel \revf{standard} deviation.}
	\label{fig:euroc_stats}
\end{figure}

\begin{figure}[h!]
	\centering
	\subfloat{\includegraphics[width=\columnwidth]{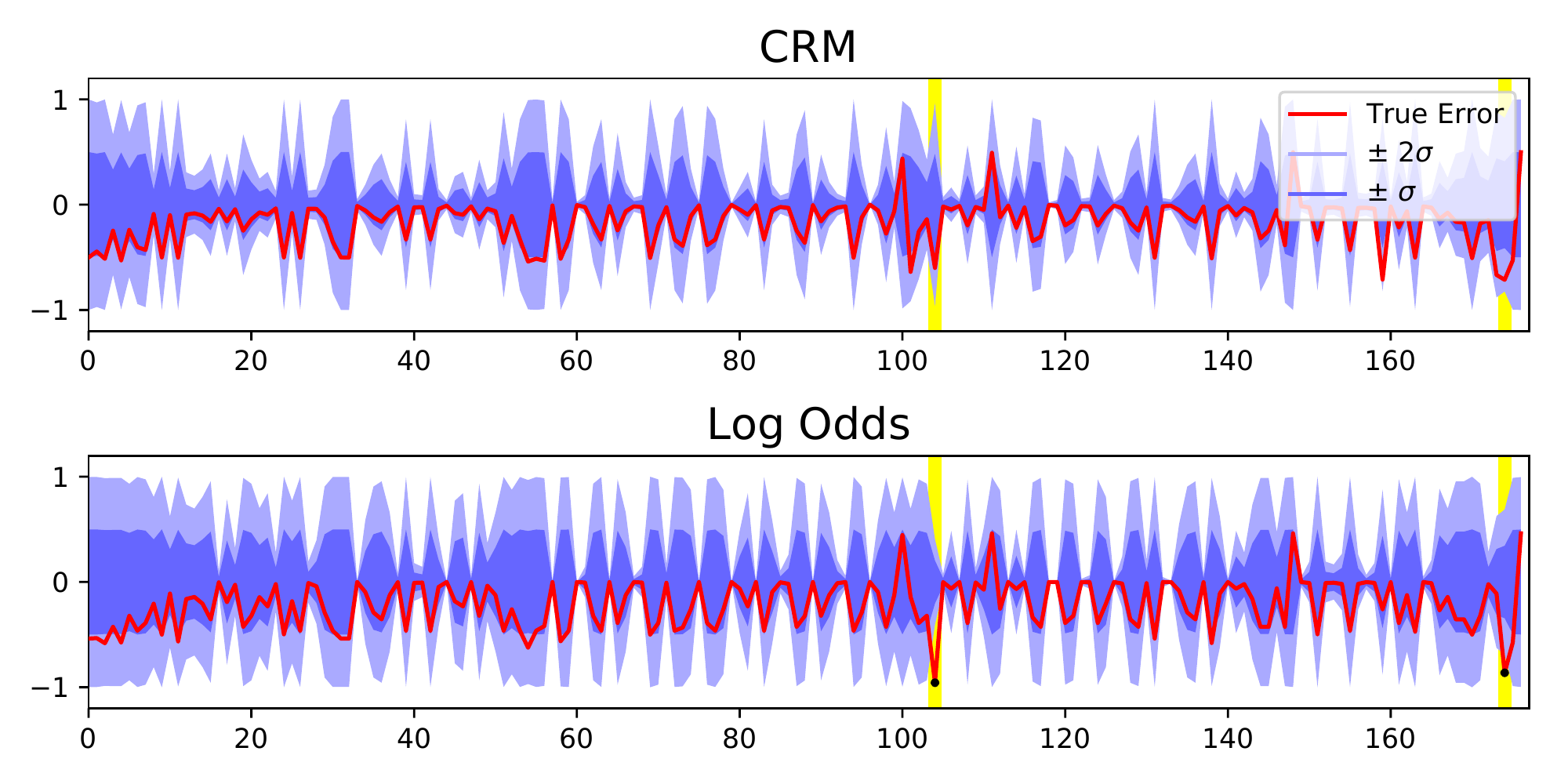}}
	\caption{True mapping error (red) and estimated \revf{standard} deviation (blue shades for $ 2\sigma $ (light) and $ \sigma $ (dark) confidence bounds) over a set of voxels for the EuRoC MAV dataset, with inconsistencies (black dots, yellow background for reference in both sub-plots) where the error is outside the $2\sigma$ confidence bound. \revf{The abscissa corresponds to the voxel id.}}
	\label{fig:euroc_voxels}
\end{figure}

We use a voxel size of \SI{6.25}{\cm} and process every 100th pixel of every 5th depth image computed from the stereo camera input (pairs of $752\times480$ monochromatic images). Fig.~\ref{fig:euroc_maps} shows the comparison results visually for mapping error, \revf{standard} deviation, and consistency.

Besides the accuracy being considerably higher in CRM compared to Log-Odds, with a mean absolute error reduction by about $0.2$ over $1,516,536$ voxels in total, the Pearson correlation coefficient (Fig.~\ref{fig:euroc_stats}) and most of the observed voxels (Fig.~\ref{fig:euroc_voxels}) also exhibit a higher consistency between estimated uncertainty (\revf{standard} deviation) and true absolute error.

\subsubsection{USC Intel Drone Dataset}~\\
We collected our own real-robot dataset using the Intel Aero quadrotor drone platform. We recorded depth measurements from a Hokuyo URG-04LX-UG01 laser scanner (as the ground-truth measurement) and from an Intel RealSense R200 imaging system (as the main perception modality for mapping) that captures depth images using an infrared stereo camera. The quadrotor's motion was not autonomous during the data collection phase.
\begin{figure}[h!]
	\centering
	\includegraphics[height=3.1cm,trim=200pt 0 0 0,clip]{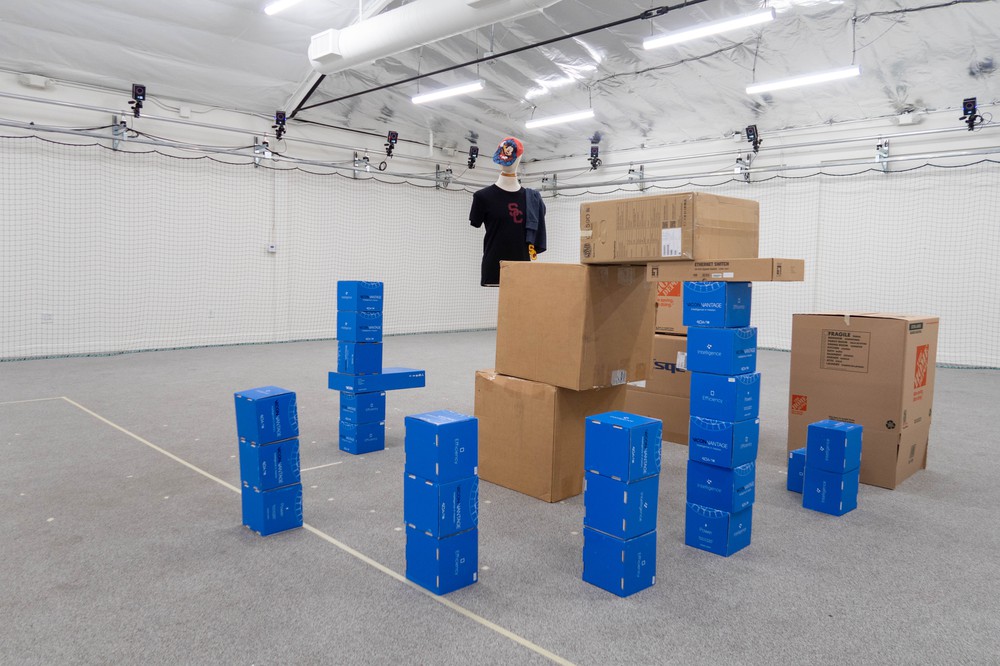}
	\includegraphics[height=3.1cm,trim=50pt 0 0 0,clip]{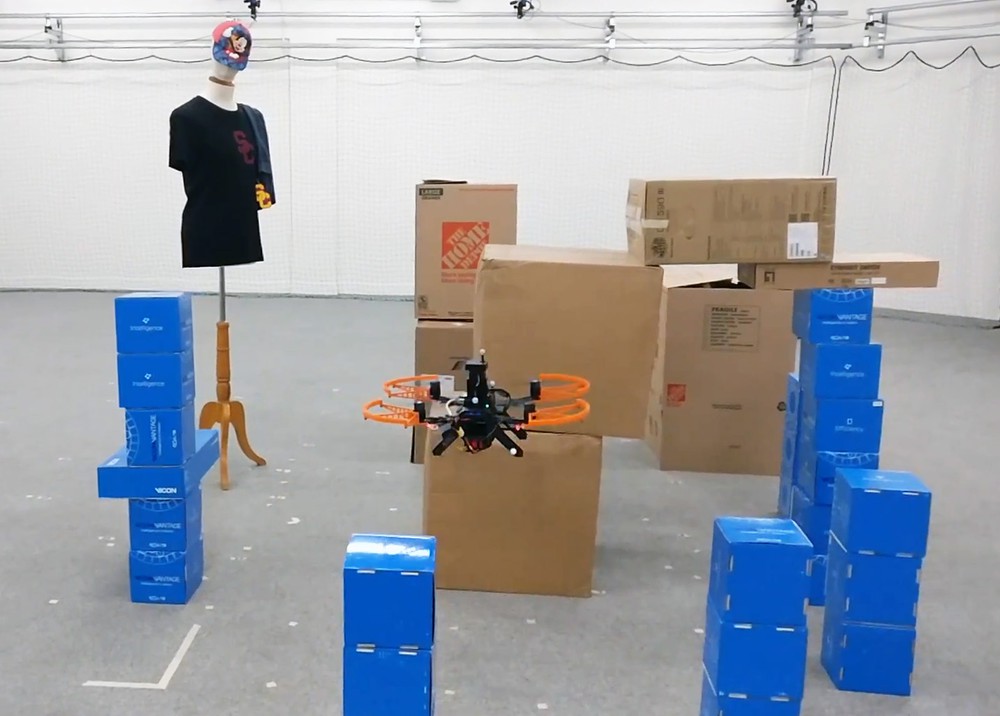} \\
	\includegraphics[width=.47\columnwidth]{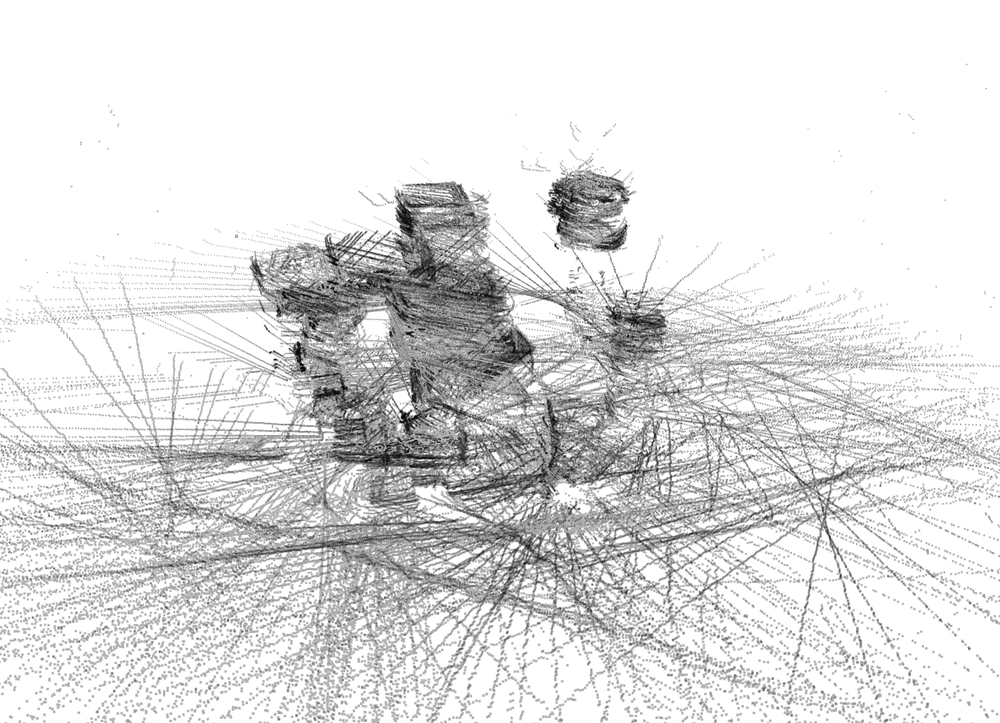}
	\includegraphics[width=.49\columnwidth]{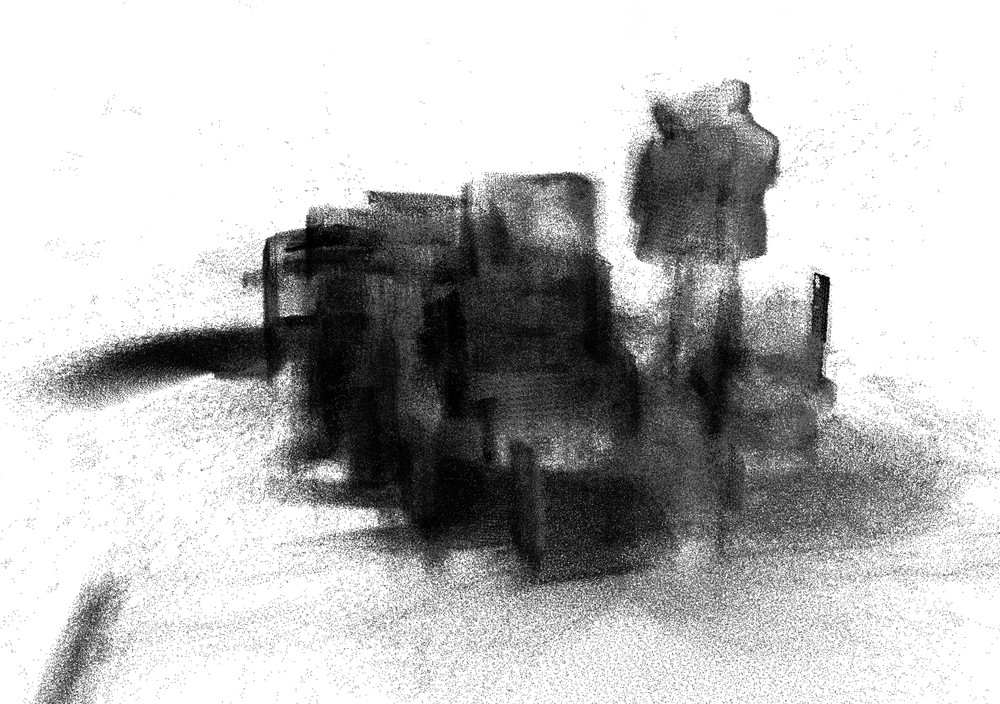}
	\caption{\textit{Top row:} Setup of the environment for the USC dataset with the \textsc{VICON} motion capture system in the background (left) and Intel Aero drone in the foreground (right).
	\textit{Bottom row:} Point cloud of laser scans taken from the (left) Hokuyo range finder for ground-truth and (right) stereo camera for mapping.}
	\label{fig:usc_setup}
\end{figure}

\begin{figure}
	\centering
	\subfloat{\includegraphics[width=.95\textwidth]{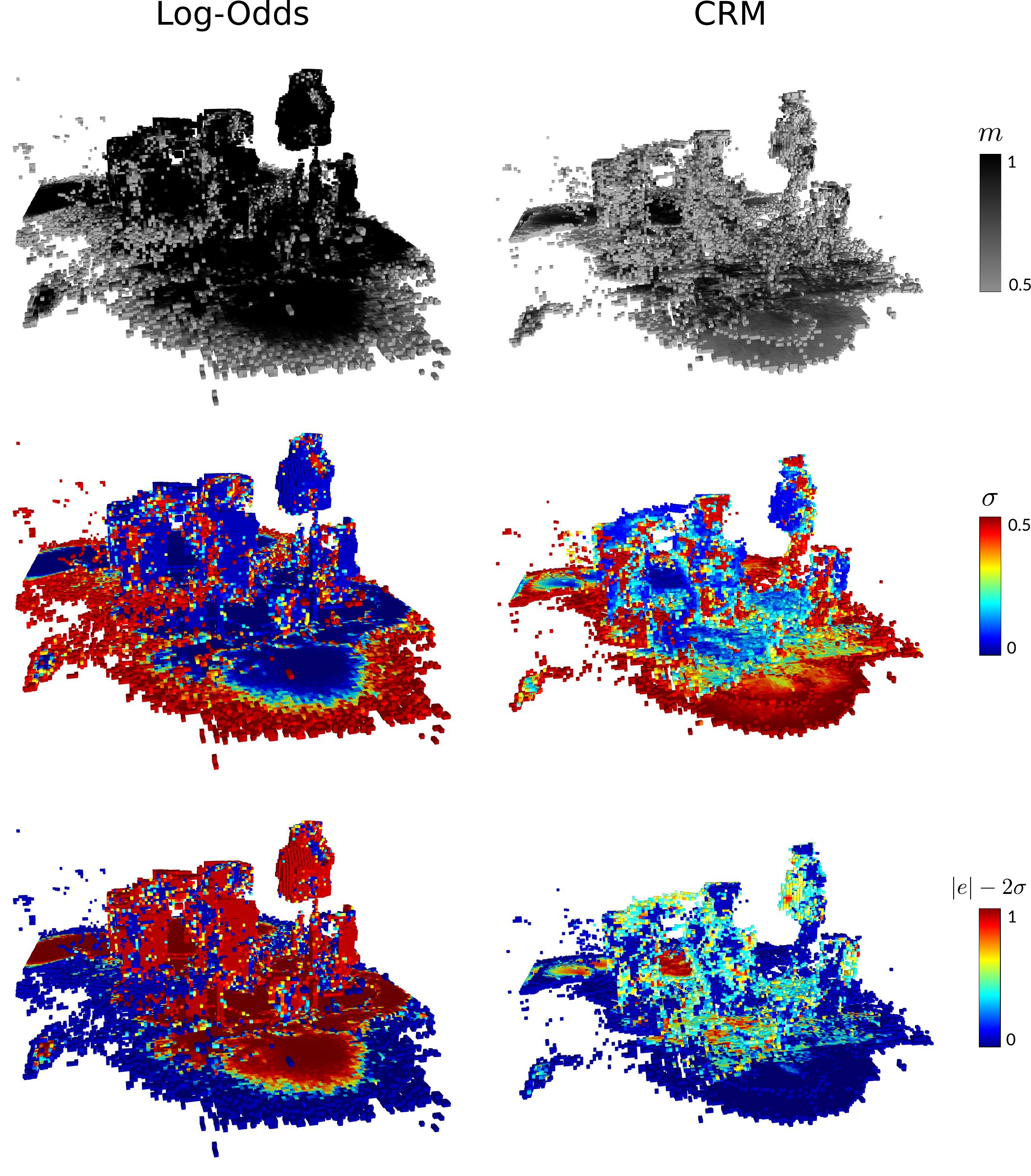}}
	\caption{\rev{Occupancy grids computed via Log-Odds (left column) and CRM (right column) resulting from the USC Intel Drone dataset. We compare two maps in terms of the voxel mean occupancy (top row), estimated \revf{standard} deviation (middle row) and inconsistency $I_c$ (bottom row) with $\gamma=2$ (Eq.~\ref{eq:inconsistencyMeasure}).} Only voxels with an estimated occupancy greater than $0.5$ are shown. Due to misaligned point clouds, the mannequin and its stand cannot be clearly represented in the map. The Log-Odds map, however, erroneously exhibits high confidence in two possible locations for the stand, leading to inconsistencies that could cause the generation of unsafe trajectories. In contrast, CRM estimates consistent and more accurate levels of confidence around the stand and other parts of the map where conflicting measurements have been acquired.}
	\label{fig:usc_maps}
\end{figure}

This experiment was conducted at the University of Southern California in a motion capture space using a \textsc{VICON} Vantage system with 24 cameras. We set up an environment where boxes of various sizes form a narrow arc and obstacle-laden passages (Fig.~\ref{fig:usc_setup}). The environment additionally features a mannequin attached to a cylindrical stand with a diameter of \SI{5}{\cm}. Due to imperfect calibration of the camera transformation with respect to the drone's body reference frame, the point-clouds obtained from the RealSense have a misalignment which poses significant challenges in mapping the mannequin as it was observed from many different viewpoints (cf.~Fig.~\ref{fig:usc_maps}).
We use the motion capture system to accurately localize the drone and compute a binary ground-truth map from the laser scans.

We compared the Log-Odds-based map and CRM computed from the stereo camera data with a voxel size of \SI{3.125}{\cm} and observed significant inconsistencies in the Log-Odds map (Fig.~\ref{fig:usc_maps}, top right), leading to an overall inconsistency $I_c$ (with $\gamma=2$) (Eq.~\ref{eq:inconsistencyMeasure}) of $6,344.294$ for Log-Odds and $1,196.418$ for CRM. Misaligned delicate structures such as the mannequin stand appeared at two different locations, with high confidence values assigned to the voxels at both places. In contrast, the richer voxel representation of CRM induces a volume of voxels spanning over the possible locations of the stand with a significantly lower confidence. This is an important feature for a representation that functions as basis for planning purposes since it enables the planner to accurately assess the collision risk and plan accordingly.

\section{Conclusion} \label{sec:conclusion}
In this work, we propose a novel algorithm for occupancy grid mapping, by storing richer data for every voxel. 
It extends traditional grid mapping in three ways: first, it maintains the probabilistic dependence between voxels within the same measurement cone. 
Second, it relaxes the need for hand-engineering an inverse sensor model and proposes the concept of ``sensor cause model" that is environment-agnostic and can be derived in a principled manner from the forward sensor model. 
Third, and most importantly, it provides consistent confidence values over the occupancy estimation that can be reliably used in planning. 
The method runs online as measurements are received and it enables mapping environments where voxels might be partially occupied.
Extensive real-world experiments show that the mapping accuracy is up to three times higher than the ISM-based method and, in simulation, $30\%$ better than Gaussian Processes maps. More importantly, according to the proposed consistency measure, the confidence values are up to two orders of magnitude more reliable.

\rev{For the future work, we plan to extend this approach in multiple directions. Tight integration with uncertainty-aware motion planners (\eg \cite{Ali14-IJRR}, \cite{SLAP_TRO18}, and \cite{Sung2019ral}) is our first goal. \cite{heiden2017planning} is a step in this direction that plans motions for quadrotors using CRM-based collision risk prediction. Second, we plan to study the extension of the method to non-Lambertian surfaces and more general light source models, where additional surface orientation information \revf{needs} to be included. Experimenting with various ranging sensors (\eg \cite{heiden2018fusion}) is a step in this direction. We will study how CRM can be extended to CRLM (confidence-rich localization and mapping) and potentially enhance grid-based SLAM solvers. Finally, we will investigate the application of this method to different domains, in particular, to perception-aware planetary exploration \cite{Kyon2018RAL,Nilsson18RSS}.}

\begin{acks}
\rev{The authors would like to acknowledge the anonymous reviewers who provided numerous suggestions and constructive criticism during the reviewing process.
We would like to thank Kiran Somasundaram, Shayegan Omidshafiei, Christopher Lott, Bardia Behabadi, Casimir Wierzynski, Sarah Paige Gibson for fruitful discussions at the preliminary stages of formalizing CRM.
The authors also thank Pradyumna Vyshnav, Daniel Pastor and James A. Preiss for their support with the experiment carried out at the University of Southern California.}
\end{acks}

\renewcommand*{\bibfont}{\footnotesize}
\bibliographystyle{SageH}
\bibliography{AliAgha.bib}

\begin{thebibliography}{43}
\providecommand{\natexlab}[1]{#1}
\providecommand{\url}[1]{\texttt{#1}}
\providecommand{\urlprefix}{URL }
\expandafter\ifx\csname urlstyle\endcsname\relax
  \providecommand{\doi}[1]{DOI:\discretionary{}{}{}#1}\else
  \providecommand{\doi}{DOI:\discretionary{}{}{}\begingroup
  \urlstyle{rm}\Url}\fi

\bibitem[{{Agha-mohammadi} et~al.(2018){Agha-mohammadi}, Agarwal, Kim,
  Chakravorty and Amato}]{SLAP_TRO18}
{Agha-mohammadi} A, Agarwal S, Kim S, Chakravorty S and Amato N (2018) {SLAP}:
  Simultaneous localization and planning under uncertainty via dynamic
  replanning in belief space.
\newblock \emph{IEEE Transactions on Robotics} 34(5): 1195--1214.

\bibitem[{{Agha-mohammadi} et~al.(2014){Agha-mohammadi}, Chakravorty and
  Amato}]{Ali14-IJRR}
{Agha-mohammadi} A, Chakravorty S and Amato N (2014) {FIRM}: Sampling-based
  feedback motion planning under motion uncertainty and imperfect measurements.
\newblock \emph{International Journal of Robotics Research (IJRR)} 33(2):
  268--304.

\bibitem[{Agha{-}mohammadi et~al.(2017)Agha{-}mohammadi, Heiden, Hausman and
  Sukhatme}]{agha2017crm}
Agha{-}mohammadi A, Heiden E, Hausman K and Sukhatme GS (2017) Confidence-rich
  grid mapping.
\newblock In: \emph{Proceedings of International Symposium of Robotics Research
  (ISRR)}.

\bibitem[{Borenstein and Koren(1991)}]{borenstein1991real}
Borenstein J and Koren Y (1991) Real-time map building for fast mobile robot
  obstacle avoidance.
\newblock In: \emph{Mobile Robots V}, volume 1388. International Society for
  Optics and Photonics, pp. 74--82.

\bibitem[{Burri et~al.(2016)Burri, Nikolic, Gohl, Schneider, Rehder, Omari,
  Achtelik and Siegwart}]{Burri25012016}
Burri M, Nikolic J, Gohl P, Schneider T, Rehder J, Omari S, Achtelik MW and
  Siegwart R (2016) The euroc micro aerial vehicle datasets.
\newblock \emph{The International Journal of Robotics Research}
  \doi{10.1177/0278364915620033}.
\newblock
  \urlprefix\url{http://ijr.sagepub.com/content/early/2016/01/21/0278364915620033.abstract}.

\bibitem[{Elfes(1989)}]{Elfes1989occupancy}
Elfes A (1989) \emph{Occupancy grids: A probabilistic framework for robot
  perception and navigation}.
\newblock PhD Thesis, Carnegie Mellon University.

\bibitem[{H{\"a}hnel et~al.(2003)H{\"a}hnel, Triebel, Burgard and
  Thrun}]{hahnel2003map}
H{\"a}hnel D, Triebel R, Burgard W and Thrun S (2003) Map building with mobile
  robots in dynamic environments.
\newblock In: \emph{IEEE International Conference on Robotics and Automation},
  volume~2. IEEE, pp. 1557--1563.

\bibitem[{Handa et~al.(2014)Handa, Whelan, McDonald and
  Davison}]{handa2014iclnuim}
Handa A, Whelan T, McDonald J and Davison AJ (2014) A benchmark for rgb-d
  visual odometry, 3d reconstruction and slam.
\newblock In: \emph{2014 IEEE International Conference on Robotics and
  Automation (ICRA)}. pp. 1524--1531.
\newblock \doi{10.1109/ICRA.2014.6907054}.

\bibitem[{Hartley and Zisserman(2000)}]{Hartley2000}
Hartley R and Zisserman A (2000) \emph{Multiple View Geometry in computer
  vision}.
\newblock Cambridge Press.

\bibitem[{Heiden et~al.(2017)Heiden, Hausman, Sukhatme and
  Agha{-}mohammadi}]{heiden2017planning}
Heiden E, Hausman K, Sukhatme GS and Agha{-}mohammadi A (2017) Planning
  high-speed safe trajectories in confidence-rich maps.
\newblock In: \emph{International Conference on Intelligent Robots and Systems,
  {IROS}}. IEEE/RSJ, pp. 2880--2886.
\newblock \doi{10.1109/IROS.2017.8206120}.
\newblock \urlprefix\url{https://doi.org/10.1109/IROS.2017.8206120}.

\bibitem[{Heiden et~al.(2018)Heiden, Pastor, Vyshnav and
  Agha{-}mohammadi}]{heiden2018fusion}
Heiden E, Pastor D, Vyshnav P and Agha{-}mohammadi A (2018) Heterogeneous
  sensor fusion via confidence-rich 3d grid mapping: Application to physical
  robots.
\newblock In: \emph{Proceedings of International Symposium on Experimental
  Robotics (ISER)}.

\bibitem[{Howard and Kitchen(1996)}]{howard1996generating}
Howard A and Kitchen L (1996) Generating sonar maps in highly specular
  environments.
\newblock In: \emph{In Proceedings of the Fourth International Conference on
  Control Automation Robotics and Vision}.

\bibitem[{Howard and Roy(2003)}]{Radish}
Howard A and Roy N (2003) The robotics data set repository (radish).
\newblock \urlprefix\url{http://radish.sourceforge.net/}.

\bibitem[{Kay(1993)}]{kay1993fundamentals}
Kay SM (1993) \emph{Fundamentals of Statistical Signal Processing, Volume I:
  Estimation Theory}.
\newblock Prentice Hall.

\bibitem[{Kim and Kim(2013)}]{kim2013occupancy}
Kim S and Kim J (2013) Occupancy mapping and surface reconstruction using local
  gaussian processes with kinect sensors.
\newblock \emph{Transactions on Cybernetics} 43(5): 1335--1346.

\bibitem[{Kim et~al.(2014)Kim, Kim et~al.}]{kim2014recursive}
Kim S, Kim J et~al. (2014) Recursive bayesian updates for occupancy mapping and
  surface reconstruction.
\newblock In: \emph{Proceedings of the Australasian Conference on Robotics and
  Automation}.

\bibitem[{Kim et~al.(2019)Kim, Thakker and {Agha-mohammadi}}]{Sung2019ral}
Kim SK, Thakker R and {Agha-mohammadi} A (2019) Bi-directional value learning
  for risk-aware planning under uncertainty.
\newblock \emph{IEEE Robotics and Automation Letters (RA-L)} In Press.

\bibitem[{Konolige(1997)}]{konolige1997improved}
Konolige K (1997) Improved occupancy grids for map building.
\newblock \emph{Autonomous Robots} 4(4): 351--367.

\bibitem[{Konolige et~al.(2008)Konolige, Agrawal, Bolles, Cowan, Fischler and
  Gerkey}]{konolige2008outdoor}
Konolige K, Agrawal M, Bolles RC, Cowan C, Fischler M and Gerkey B (2008)
  Outdoor mapping and navigation using stereo vision.
\newblock In: \emph{Experimental Robotics}. Springer, pp. 179--190.

\bibitem[{Kutulakos and Seitz(2000)}]{kutulakos2000shape}
Kutulakos KN and Seitz SM (2000) A theory of shape by space carving.
\newblock \emph{International Journal of Computer Vision} 38(3): 199--218.
\newblock \doi{10.1023/A:1008191222954}.
\newblock \urlprefix\url{https://doi.org/10.1023/A:1008191222954}.

\bibitem[{Martin and Aggarwal(1983)}]{martin1983volumetric}
Martin WN and Aggarwal JK (1983) Volumetric descriptions of objects from
  multiple views.
\newblock \emph{IEEE transactions on pattern analysis and machine intelligence}
  (2): 150--158.

\bibitem[{Moravec(1988)}]{moravec1988sensor}
Moravec HP (1988) Sensor fusion in certainty grids for mobile robots.
\newblock \emph{AI magazine} 9(2): 61.

\bibitem[{Moravec(1996)}]{moravec1996TR}
Moravec HP (1996) Robot spatial perception by stereoscopic vision and 3d
  evidence grids.
\newblock \emph{Technical Report CMU-RI-TR-96-34, Carnegie Mellon University} .

\bibitem[{Newcombe et~al.(2011)Newcombe, Izadi, Hilliges, Molyneaux, Kim,
  Davison, Kohi, Shotton, Hodges and Fitzgibbon}]{newcombe2011kinectfusion}
Newcombe RA, Izadi S, Hilliges O, Molyneaux D, Kim D, Davison AJ, Kohi P,
  Shotton J, Hodges S and Fitzgibbon A (2011) Kinectfusion: Real-time dense
  surface mapping and tracking.
\newblock In: \emph{Mixed and augmented reality (ISMAR), 2011 10th IEEE
  international symposium on}. IEEE, pp. 127--136.

\bibitem[{Nilsson et~al.(2018)Nilsson, Haesaert, Thakker, Otsu, Vasile,
  {Agha-mohammadi}, Murray and Ames}]{Nilsson18RSS}
Nilsson P, Haesaert S, Thakker R, Otsu K, Vasile CI, {Agha-mohammadi} A, Murray
  R and Ames A (2018) Toward specification-guided active mars exploration for
  cooperative robot teams.
\newblock Pittsburgh, Pennsylvania.
\newblock \doi{10.15607/RSS.2018.XIV.047}.

\bibitem[{O'Callaghan et~al.(2009)O'Callaghan, Ramos and
  Durrant-Whyte}]{OCallaghan2009contextual}
O'Callaghan S, Ramos FT and Durrant-Whyte H (2009) Contextual occupancy maps
  using gaussian processes.
\newblock In: \emph{International Conference on Robotics and Automation}. IEEE,
  pp. 1054--1060.

\bibitem[{O'Callaghan and Ramos(2012)}]{OCallaghan2012-IJRR}
O'Callaghan ST and Ramos FT (2012) Gaussian process occupancy maps.
\newblock \emph{The International Journal of Robotics Research} 31(1): 42--62.

\bibitem[{Oleynikova et~al.(2017)Oleynikova, Taylor, Fehr, Siegwart and
  Nieto}]{oleynikova2017voxblox}
Oleynikova H, Taylor Z, Fehr M, Siegwart R and Nieto J (2017) Voxblox:
  Incremental 3d euclidean signed distance fields for on-board mav planning.
\newblock In: \emph{IEEE/RSJ International Conference on Intelligent Robots and
  Systems (IROS)}.

\bibitem[{Otsu et~al.(2018)Otsu, {Agha-mohammadi} and Paton}]{Kyon2018RAL}
Otsu K, {Agha-mohammadi} A and Paton M (2018) Where to look? predictive
  perception with applications to planetary exploration.
\newblock \emph{IEEE Robotics and Automation Letters (RA-L)} 3(2): 635--642.

\bibitem[{Pagac et~al.(1996)Pagac, Nebot and
  Durrant-Whyte}]{pagac1996evidential}
Pagac D, Nebot EM and Durrant-Whyte H (1996) An evidential approach to
  probabilistic map-building.
\newblock In: \emph{Reasoning with Uncertainty in Robotics}. Springer, pp.
  164--170.

\bibitem[{Paskin and Thrun(2005)}]{Paskin05}
Paskin M and Thrun S (2005) Robotic mapping with polygonal random fields.
\newblock In: \emph{Conference on Uncertainty in Artificial Intelligence}. p.
  450–458.

\bibitem[{Ramos and Ott(2016)}]{ramos2016hilbert}
Ramos F and Ott L (2016) Hilbert maps: scalable continuous occupancy mapping
  with stochastic gradient descent.
\newblock \emph{The International Journal of Robotics Research} 35(14):
  1717--1730.

\bibitem[{Schaefer et~al.(2018)Schaefer, Luft and Burgard}]{schaefer2018dct}
Schaefer A, Luft L and Burgard W (2018) Dct maps: Compact differentiable lidar
  maps based on the cosine transform.
\newblock \emph{IEEE Robotics and Automation Letters} 3(2): 1002--1009.
\newblock \doi{10.1109/LRA.2018.2794602}.

\bibitem[{Senanayake et~al.(2017)Senanayake, O'Callaghan and
  Ramos}]{senanayake2017learning}
Senanayake R, O'Callaghan S and Ramos F (2017) Learning highly dynamic
  environments with stochastic variational inference.
\newblock In: \emph{Robotics and Automation (ICRA), 2017 IEEE International
  Conference on}. IEEE, pp. 2532--2539.

\bibitem[{Stachniss(2009)}]{stachniss2009_book}
Stachniss C (2009) \emph{Robotic mapping and exploration}, volume~55.
\newblock Springer.

\bibitem[{Thrun(1998)}]{thrun1998learning}
Thrun S (1998) Learning metric-topological maps for indoor mobile robot
  navigation.
\newblock \emph{Artificial Intelligence} 99(1): 21--71.

\bibitem[{Thrun(2003)}]{thrun2003learning}
Thrun S (2003) Learning occupancy grid maps with forward sensor models.
\newblock \emph{Autonomous robots} 15(2): 111--127.

\bibitem[{Thrun et~al.(2005)Thrun, Burgard and Fox}]{Thrun2005}
Thrun S, Burgard W and Fox D (2005) \emph{Probabilistic Robotics}.
\newblock MIT Press, Cambridge, MA.

\bibitem[{Thrun et~al.(2002)}]{thrun2002robotic}
Thrun S et~al. (2002) Robotic mapping: A survey.
\newblock \emph{Exploring artificial intelligence in the new millennium} 1:
  1--35.

\bibitem[{Veeck and Burgard(2004)}]{veeck2004learning}
Veeck M and Burgard W (2004) Learning polyline maps from range scan data
  acquired with mobile robots.
\newblock In: \emph{IEEE/RSJ International Conference on Intelligent Robots and
  Systems}, volume~2. pp. 1065--1070.

\bibitem[{Wang and Englot(2016)}]{wang2016fast}
Wang J and Englot B (2016) Fast, accurate gaussian process occupancy maps via
  test-data octrees and nested bayesian fusion.
\newblock In: \emph{IEEE International Conference on Robotics and Automation
  (ICRA)}. pp. 1003--1010.

\bibitem[{Wurm et~al.(2010)Wurm, Hornung, Bennewitz, Stachniss and
  Burgard}]{wurm2010octomap}
Wurm KM, Hornung A, Bennewitz M, Stachniss C and Burgard W (2010) Octomap: A
  probabilistic, flexible, and compact 3d map representation for robotic
  systems.
\newblock In: \emph{ICRA 2010 workshop on best practice in 3D perception and
  modeling for mobile manipulation}, volume~2.

\bibitem[{Yamauchi(1997)}]{yamauchi1997frontier}
Yamauchi B (1997) A frontier-based approach for autonomous exploration.
\newblock In: \emph{IEEE International Symposium on Computational Intelligence
  in Robotics and Automation}. pp. 146--151.

\end{thebibliography}

\end{document}